%% file: main.tex
\documentclass{article}
\usepackage{amssymb}
\usepackage{amsthm}
\usepackage{physics}
\usepackage{commands}

\usepackage{graphicx}
\graphicspath{{figs/}}
\usepackage{subfigure}
\usepackage{import}

\usepackage{mathrsfs}
\usepackage{bbm}
\usepackage{libertine} 
\usepackage[numbers]{natbib}

\usepackage{enumitem}

\usepackage{xcolor}
\usepackage[normalem]{ulem}
\usepackage[margin=1in]{geometry}

\newcommand{\checked}[1]{\textcolor{black}{#1}}

\newcommand{\cossim}{\mathrm{cossim}}

\begin{document}

\title{The Effect of SGD Batch Size on Autoencoder Learning: \\Sparsity, Sharpness, and Feature Learning}
\author{Nikhil Ghosh\\
Department of Statistics\\University of California, Berkeley\\
\texttt{\sf nikhil\_ghosh@berkeley.edu}
\and
Spencer Frei\\
Simons Institute for the Theory of Computing\\
University of California, Berkeley\\
\texttt{\sf frei@berkeley.edu}
\and
\hspace{-2em}
Wooseok Ha\\
\hspace{-2em} Department of Statistics\\
\hspace{-2em}  University of California, Berkeley\\
\hspace{-2em}  \texttt{\sf haywse@berkeley.edu}
\and
\hspace{1.5em} Bin Yu\\
\hspace{1.5em} Department of Statistics and EECS\\ 
\hspace{1.5em} University of California, Berkeley\\
\hspace{1.5em} \texttt{\sf binyu@stat.berkeley.edu}
}

\maketitle

\begin{abstract}
\subimport{sections/}{abstract.tex}
\end{abstract}

\section{Introduction}
\subimport{sections/}{introduction.tex}

\section{Related Work}\label{sec:related}
\subimport{sections/}{related.tex}
\subimport{sections/}{results.tex}
\section{Conclusion}\label{sec:conclusion}
\subimport{sections/}{conclusion.tex}

\section*{Acknowledgements}
This research is kindly supported in part by the NSF and the Simons Foundation for the Collaboration on the Theoretical Foundations of Deep Learning through awards DMS-2031883 and 814639. NG would 
like to acknowledge support from the NSF RTG Grant \#1745640, and BY would like to acknowledge partial support from the NSF Grant DMS-2015341. WH was supported from NSF TRIPODS Grant 1740855, DMS-1613002, and 2015341.


\vskip 0.2in
\bibliographystyle{plainnat}
\bibliography{refs}
\newpage

\appendix
\subimport{sections/}{appendix.tex}

\end{document}

%% file: sections/abstract.tex
In this work we investigate the dynamics of stochastic gradient descent (SGD) when training a single-neuron autoencoder with \checked{linear or ReLU activation on orthogonal data.} We show that for this non-convex problem, 
randomly initialized SGD with a constant step size successfully finds a global minimum for any batch size choice.
However, the particular global minimum found depends upon the batch size.  In the full-batch setting, 
we show that the solution is dense (i.e., not sparse) and is highly aligned with its initialized direction, showing that relatively little feature learning occurs. 
On the other hand, for any batch size strictly smaller than the number of samples, SGD finds a global minimum which is sparse and nearly orthogonal to its initialization, showing that the randomness of stochastic gradients induces a qualitatively different type of ``feature selection'' in this setting. 
Moreover, if we measure the sharpness of the minimum by the trace of the Hessian, the minima found with full batch gradient descent are \textit{flatter} than those found with strictly smaller batch sizes, in contrast to previous works which suggest that large batches lead to sharper minima.   
To prove convergence of SGD with a constant step size, we introduce a powerful tool from the theory of non-homogeneous random walks which may be of independent interest.

%% file: sections/introduction.tex
Recent years have witnessed impressive successes of neural networks across a wide variety of domains. However, their ability to generalize to unseen data is still not fully understood \citep{zhang2021understanding, neyshabur2017exploring}. \checked{One potential explanation is that gradient-based optimization algorithms have an ``implicit bias" towards 
particular solutions which have simple structure, e.g. small norm or rank~\citep{vardi2022implicit,ji2020directional,azulay2021implicit,gunasekar2018characterizing,gunasekar2018implicit,soudry2018implicit,neyshabur2017geometry,boursier2022gradient,frei2022implicit}.  In some instances, these solutions provably achieve small generalization error~\citep{woodworth2020kernel,safran2022relulinear,frei2023sword}.}  It has been observed that the choice of step size and batch size in these algorithms can make a substantial difference in the generalization performance of trained neural networks, with generally better performance obtained when using larger step sizes and smaller batch sizes~\citep{keskar2016large,jastrzkebski2017three,wu2018sgd}. 

These observations have inspired a surge of research aimed at more deeply \checked{understanding the particular effects of small batch size~\citep{haochen2021shape,damian2021label,mulayoff2020unique,wu2022does} and step-size~\citep{li2019largelearningrate,even2023s} on SGD training.  }
However, most prior theoretical works have not directly analyzed the effect of \textit{mini-batch noise}. Instead these works model the gradient noise by considering algorithms which explicitly add noise to the labels or to the gradients, often further assuming a vanishingly small step size~\citep{haochen2021shape,damian2021label,xie2021a,li2022sgdzeroloss}. This motivates the main questions we investigate in this work: 
\begin{center}
\textit{Are there settings where \textbf{standard} SGD training of a neural network converges to qualitatively different solutions based on the choice of batch size?}
\end{center}

To investigate this question, we consider the setting of a single neuron autoencoder with either a linear or ReLU activation trained on orthonormal data. One can view the data as arising from a special case of a sparse coding model where the dictionary is orthogonal and the latent codes are one-hot vectors (see Section~\ref{sec:related} for more details). To be concrete, we will refer to any training algorithm which updates the parameters by subtracting a multiple of the gradient of a mini-batch as a \textbf{mini-batch GD} algorithm. Let $m$ denote the total number of training points. From now on, we will use the following naming conventions
\begin{itemize}
    \item \textbf{Mini-batch SGD} (or just SGD) will refer to the mini-batch GD algorithm where at each iteration a mini-batch of size $b < m$ is drawn uniformly with replacement.

    \item \textbf{Full-batch GD} (or just GD) will refer to the mini-batch GD algorithm where at each iteration the mini-batch is just the full dataset of $m$ points.
\end{itemize}
Although the model and data we consider are stylized, the setting is rich enough to allow for us to probe the effect of \textit{mini-batch noise} in SGD in a  \textit{non-convex} setting, for which there exists a whole \textit{manifold} of global minima. In particular, we show that in this setting there are a number of striking differences in the solutions found by SGD in comparison to GD. We will say that a solution is ``sparse'' if it can be expressed as a sparse linear combination of the data, which serves to further highlight the connection with the sparse coding model mentioned earlier. We now summarize \checked{at a high level} our main contributions \checked{and their implications, which hold for both linear and ReLU activations}. 
\begin{enumerate}
    
     \item In the full-batch setting, \checked{randomly initialized} GD converges to a \textit{dense} global minimum that is a \checked{nearly uniform mixture of many training data points}. This minimum is just a rescaling of the initialization projected onto \checked{a subset of} the span of the data. 
    
    \item For any batch size strictly smaller than \checked{the size of the dataset}, SGD converges \checked{almost surely} to 
    \checked{a single datapoint, which is a 1-\textit{sparse} global minimum that is nearly orthogonal to the random initialization.} \checked{Notably, the SGD convergence result holds for a constant step size and any batch size strictly smaller than the number of samples.  } 

    \item  \checked{We show that}
    GD exhibits relatively little feature learning since the learned solution is in nearly the same direction as its random initialization, whereas the SGD solution is nearly orthogonal to it. Additionally, the GD solution is invariant to certain orthogonal transformations of the data while the SGD solution is not, further illustrating that SGD learns a more data dependent solution.
    
    \item If we measure the sharpness of the solution found by the trace of the Hessian, we show that SGD converges to \textit{sharper} minima than GD when the activation is ReLU. In contrast, previous works hypothesize that smaller batches result in flatter minima~\citep{keskar2016large}, which suggests a potential weakness of this measure of sharpness. 
\end{enumerate}

Our results hold by a careful analysis of the trajectory of SGD/GD following a standard random initialization scheme. We show that for orthonormal data, the ReLU autoencoder dynamics reduce to that of a linear autoencoder trained on a subset of the data. Thus it suffices for us to analyse the linear autoencoder dynamics. The loss landscape of linear autoencoders has been studied in the past~\citep{baldi1989linearautoencoderpca,plaut2018linearautoencoderpca, kunin2019loss}, as well as their gradient dynamics~\citep{gidel2019implicit, bao2020regularized} which are closely related to Oja's rule from neuroscience~\citep{oja1982simplified, yan1994global} and the streaming PCA problem~\citep{shamir2016convergence, allen2017first}. However, no prior work has studied the convergence of gradient methods in parameter space when the first principal component is not unique, as is the case in our setting; nor has prior work highlighted the role of batch size. In particular, the SGD case requires significant technical innovation as we are considering the dynamics under a constant (fixed) step size and we cannot couple the SGD trajectory with that of GD since
they converge to qualitatively different minima. 

Indeed, most classical analyses of SGD (e.g., \cite{robbins1951stochastic}) assume the step size decays to zero as this is what is generally required to ensure that the iterates can converge to single point rather than to a measure with full support. However, it is not always required to decay the step size for training to converge. 
Notably,~\citet{nacson19sgdseparabledata} proved that for linearly separable data, a linear model trained with SGD on the logistic loss with a constant step size converges in direction to the $\ell_2$-max-margin predictor. This however is identical to the convergence behavior of full-batch GD~\citep{soudry2018implicit}, hence it is impossible to isolate the effect of stochastic gradients on the types of solutions found by SGD/GD in this setting. \checked{In practice, constant step size SGD often suffices to fully optimize the training loss and achieve competitive generalization capabilities~\citep{soudry2018implicit}, making it a practical baseline as it requires less tuning than more complicated schedules. In particular, step size decay is usually employed not to better optimize the training loss, but to improve the generalization performance.}

To analyze constant step size SGD in our setting we introduce a powerful tool from the theory of non-homogeneous random walks to develop convergence guarantees. Such tools first appeared in the probability literature~\citep{menshikov2010rate}, but to our knowledge have not been 
employed in a machine learning setting 
prior to this work. At a high level, our proof works by showing that at each iteration SGD 
amplifies
the correlation of the weights with a particular data direction relative to all other data directions. By viewing this relative correlation as a stochastic process induced by SGD, we can invoke our tool to show that the stochastic process is transient and then show that this implies convergence to a global minimum. \checked{In contrast, full-batch GD exhibits a symmetry which ensures that this process remains fixed at initialization. This symmetry is broken by the random subsampling of SGD which leads to the ``phase transition" in the asymptotic convergence behavior when the batch size changes. Note that although the limiting convergence behavior of SGD is equivalent for all batch sizes, different sizes can yield different ``rates of escape" of the associated stochastic process, which leads to different asymptotic convergence rates.}
We believe the techniques we develop for the convergence of constant step size SGD in this setting may hold wider applicability for the analysis of other machine learning algorithms. 

\checked{The rest of the paper is organized as follows. In Section~\ref{sec:related}, we provide a review of related work, and Section~\ref{sec:setting} formally presents our problem setting. We then present our theoretical results for the linear autoencoder in Section~\ref{sec:theory_lin_act}, where we provide proof sketches for the convergence of GD and SGD in Section~\ref{sec:gd_convergence} and Section~\ref{sec:sgd_convergence}, respectively. In Section~\ref{sec:theory_relu_act}, we present the corresponding convergence results for the ReLU autoencoder, and in Section~\ref{sec:loss_landscape} we compare the local loss landscapes of the ReLU autoencoder at different algorithmic solutions. Finally, we conclude in Section~\ref{sec:conclusion} where we also provide potential avenues for further work. }

%% file: sections/related.tex
\paragraph{Linear Autoencoders and Streaming PCA.} It has long been known that there is a strong connection between PCA and linear neural networks. \citet{baldi1989linearautoencoderpca} showed that a linear autoencoder with squared loss has a minimum \checked{which is the projection of the data onto the subspace} spanned by the first principal components of the training data. A variety of works actually analyze algorithms for recovering the PCA subspace. \citet{oja1982simplified} proposed a biologically plausible update rule known as Oja's rule for training a single neuron in an online setting to recover the first PCA direction. The computer science community has considered other algorithms including variations of Oja's rule for solving PCA in the streaming setting in a space efficient manner~\citep{shamir2016convergence, allen2017first, mitliagkas2013memory, jain2016streaming}. There have also been works which specifically analyze PCA recovery via training linear autoencoders using gradient descent or gradient flow~\citep{gidel2019implicit, min2021explicit, saxe2013exact}. However, all of these works only consider convergence of the loss to its minimum and not the convergence of the weights which we show can be heavily dependent on the choice of batch size. It is important to study the learned weights \checked{as we do in this work} since in practice these correspond to learned features which can have a significant impact when used for downstream tasks.

\paragraph{\checked{Neural networks and sparsity.}}
\checked{The sparse coding data model (i.e., data of the form $x = Az + \eps$, where the ``latent code" $z$ is sparse, the ``dictionary" $A$ is unknown, and $\eps$ is a noise variable) is a widely-used generative model for natural data~\citep{olshausen1997sparsecoding,vinje2000sparsecoding,olshausen2004sparsecoding}.  A number of previous works have studied different ``dictionary learning" algorithms designed to recover the hidden dictionary $A$ given data generated from the sparse coding model e.g., \cite{arora2015simple, agarwal2016learning}. More closely related to our setting is \citet{nguyen2019dynamics} which shows that two-layer autoencoders trained by variants of full-batch gradient descent on sparse coding data can \textit{locally} converge to the ground truth dictionary. In our work, we also analyze gradient descent trained autoencoders trained on a (simplified instance of) sparse coding data. However, we provide \textit{global} convergence guarantees across a range of batch sizes.   }

\paragraph{Batch size, sharpness, and generalization.} It has been empirically observed that in practice large batch size tends to degrade the generalization performance of SGD~\citep{keskar2016large} on supervised learning tasks. One contending hypothesis for this behavior is that larger batch sizes result in SGD finding ``sharper" minima which generalize poorly. Recently, there have been several theoretical works which try to make this intuition rigorous \checked{by studying SGD with explicitly added label noise}~\citep{blanc2020implicit, damian2021label, haochen2021shape, li2022sgdzeroloss}. At a high level these works show that \checked{the added}
label noise has the following implicit regularization effect: once the iterates reach a global minimum of the training loss the iterates approximately remain on the manifold of global minima, but now move to decrease a regularization term. This regularization term can be viewed as the sharpness of the \checked{loss} and is approximately equal to the trace of the Hessian for small step sizes. One can show that for certain problems such as sparse overparametrized linear regression~\citep{woodworth2020kernel} that decreasing this notion of sharpness leads to better generalization. We show that this notion of sharpness may not be universally appropriate, since in our setting smaller batch size leads to a \textit{sharper} solution. 
\checked{Moreover, in comparison to prior works~\citep{blanc2020implicit, damian2021label, haochen2021shape, li2022sgdzeroloss}, we do not require independent noise to be added to the gradient updates to model SGD noise but rather we explicitly characterize the effect of the randomness that comes from using stochastic gradients in SGD.}

\paragraph{\checked{Feature learning in neural networks.}} \checked{ Neural networks trained by gradient descent have shown a remarkable ability to learn data-dependent features which enable generalization to a variety of domains.  A number of recent works have explored how different aspects of the training procedure, including network architecture and optimization hyperparameters, affect this `feature learning' ability.  \citet{jacot2018ntk,chizat2018note} showed that when the width of neural networks grows to infinity, the learning rate is small and the random initialization has a large variance, the training dynamics of the network can be approximated by the behavior of a data-independent kernel defined by an infinite-width limit of the network at its random initialization.  In this `kernel regime' setting, the network behaves similarly to a \textit{random} feature model and no data-dependent feature learning occurs.  By contrast, when the learning rate is large and the scale of initialization is small, neural networks are indeed capable of learning data-dependent features, as has been shown in a number of recent works~\citep{wei2020regularization,allenzhu2022featurepurification,yang2022featurelearning,frei2022rfa,zou2023generalizationadam}.  However, none of these works examined how the batch size of SGD could significantly affect the types of features found by SGD, as we do in this work. In our work, we see that despite the fact that the weights move non-trivially for both GD and SGD, it turns out that SGD displays more data-dependent feature learning behavior.}

\paragraph{Dynamics of GD for single-neurons.} We note that previous works have also considered the dynamics of gradient descent for learning single-neuron architectures, e.g.,~\cite{yehudai2020learning,frei2020singleneuron,vardi2021learning,mei2018landscape}. However, none of these previous works considered unsupervised learning with autoencoders or establish a separation between the minima learned using gradient descent with different batch sizes.

\paragraph{Convergence of SGD with a fixed step-size.}  Notably, we guarantee that the final iterate of constant step-size SGD converges almost surely to a single point.   Typical convergence guarantees for SGD require either a decaying step-size, iterate averaging, or only hold in expectation or with high probability~\citep{jain2019making,sebbouh2021sure,liu2022sure, zou2021benign}.  Prior works have shown that for constant step size SGD the (averaged) iterates almost surely converge to an invariant distribution~\citep{merad2023convergence, yu2020analysis}. In general the limiting invariant distribution will have non-zero variance, however our problem has the interesting feature that the last iterate converges to a single point mass with zero variance. To the best of our knowledge, the only other such example is provided in~\cite{nacson19sgdseparabledata}.

%% file: sections/results.tex
\section{Setting}\label{sec:setting}
We consider a single neuron weight-tied auto-encoder $f: \R^n \to \R^n$ defined as
\begin{equation}\label{eq:autoencoder_definition}
    f(\bx; \bw) = \bw \phi(\ip{\bw}{\bx}).
\end{equation}
The \checked{network} 
takes as input $\bx \in \R^n$ and is parameterized by a single neuron $\bw \in \R^n$ with activation $\phi$ and no bias. We will take the activation $\phi$ to be either the identity $\phi(z)=z$ or ReLU $\phi(z) = \max(0,z)$. Furthermore, we will assume that we are given a training dataset $\dataset = \{\ba_1, \ldots, \ba_m\}$ where the $\ba_i \in \R^n$ are orthonormal and necessarily $m \leq n$. Let $(\ba_1, \ba_2, \ldots, \ba_n)$ be the completion of $\dataset$ to an orthonormal basis of $\R^n$. We will be interested in characterizing the dynamics of (stochastic) gradient descent on the standard reconstruction objective
\begin{equation}\label{eq:autoencoder_objective}
    \cL(\bw; \dataset) = \frac{1}{m }\sum\limits_{i=1}^m \ell(\bw; \ba_i),
\end{equation}
where the pointwise loss $\ell$ is the squared-loss
\[
\ell(\bw; \bx) = \frac{1}{2}\norm{\bx - f(\bx; \bw)}^2.
\]
We will consider mini-batch GD training with non-zero initialization and constant step-size $\eta > 0$, namely for $t = 0, 1, \ldots$
\begin{equation}\label{eq:minibatch_gd}
    \bw_{t+1} = \bw_t - \eta \sum_{i \in \cB_t} \grad_{\bw}~ \ell(\bw; \ba_i),~~\cB_t \subseteq [m],
\end{equation}
where the gradient of the pointwise loss $\grad_{\bw}~\ell(\bw; \bx)$ is
\begin{equation}\label{eq:grad_single_loss}
     \phi'(\sip{\bw}{\bx}) \cdot [\bx \bw^\sT + \sip{\bw}{\bx} \cdot \id_n] \cdot (f(\bx; \bw) - \bx),
\end{equation}
and we take $\phi'(t) := \mathbbm{1}(t > 0)$ when $\phi$ is ReLU. 

There are many possible instantiations of mini-batch GD training based on the mini-batch selection in Eq.\ (\ref{eq:minibatch_gd}). We will consider algorithms with fixed batch size $b := |\cB_t|$ where $b \in [m]$. In particular, we consider the following algorithms:
\begin{itemize}
    \item  Full-batch GD  where $\cB_t = [m]$ for all $t$. Note that in this case $b = m$.
    \item  Mini-batch SGD where each $\cB_t$ is chosen uniformly at random from the set of subsets of $[m]$ of size $b$ and $b < m$.
    \item  Cyclic SGD\footnote{Note that the mini-batch order is actually deterministic.} where $\cB_t = \{t \; \mathrm{mod} \; m\}$. Note that in this case $b = 1$.
\end{itemize}
We will often just refer to mini-batch SGD as SGD and full-batch GD as GD for short.  

\subsection{Visualizations of Convergence Behavior}
\begin{figure*}
    \begin{center}
    \centerline{\includegraphics[scale=1.0]{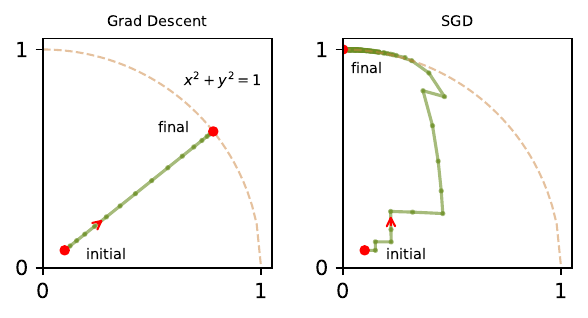}}
    \vskip -0.1in
    \caption{Visualizations of the trajectories of $\bw_t \in \R^2$ for GD and SGD (with $b=1$) when $m=n=2$. Both methods are initialized at $\bw_0=(0.1,0.08)^\top$ and run with step size $\alpha=1/4$ on the dataset $\cD = \{\ba_1, \ba_2\}$ consisting of the standard basis vectors.}
    \label{fig:trajectory}
    \end{center}
    \vskip -0.3in
    \end{figure*}
To demonstrate how the batch size influences the solutions found by gradient descent, we train a linear autoencoder using full-batch GD and stochastic GD on an example where $\dataset = \{\ba_1, \ba_2\}$ with $\ba_1=(1,0)^\top$ and $\ba_2=(0,1)^\top$. Both methods are initialized at the same point $\bw_0=(0.1,0.08)^\top$. Since the iterates $\bw_t$ lie in $\R^2$, we can visualize the optimization trajectories for each method in Figure~\ref{fig:trajectory}. In the figure, we also draw the upper quadrant of the unit circle as a dashed curve. Later in Section \ref{sec:loss_landscape} we will show that all points on the quarter circle are global minima. We see that full-batch GD converges to a point in the interior of the quarter circle, whereas SGD converges to a boundary point. As we will see in the next section, we can theoretically understand the behavior of these simulations.

\section{Theoretical Results for Linear Activation}\label{sec:theory_lin_act}
In this section we will consider the convergence behavior of the single neuron linear autoencoder define in Eq.\ (\ref{eq:autoencoder_definition}) when trained with full-batch GD, mini-batch SGD, and cyclic SGD. We will start by stating the results and then in later sections provide proof sketches of GD and SGD convergence, as well as additional relevant background. More specifically, in Section \ref{sec:gd_convergence} we give a proof sketch of GD convergence. In the remaining sections we work towards sketching the proof of SGD convergence. We start with giving some useful results about the iterates of mini-batch \checked{(S)}GD in Section \ref{sec:minibatch_gd} that are mostly algebraic. In Section \ref{sec:sgd_random_walk} we identify a stochastic process arising from SGD which is key to understanding its convergence behavior. To analyze this process, in Section \ref{sec:random_walk} we introduce a useful probabilistic result from the theory of non-homogeneous
random 
walks. Finally in Section \ref{sec:sgd_convergence} we combine all the previous results to give a proof sketch of mini-batch SGD convergence.

Let us start by introducing some notation. Given a set of indices $\cS \subseteq [n]$, we write the orthogonal projection onto $\Span(\ba_i : i \in \cS)$ as $\Pi_{\cS}$ where
\begin{equation}\label{eq:proj_pos_set}
    \Pi_{\cS}(\bx) := \sum\limits_{i \in \cS} \ip{\ba_i}{\bx} \ba_i.
\end{equation}
For convenience, define $\Pi_m := \Pi_{[m]}$. We now present our convergence result for full-batch GD. We give a sketch of the proof in Section \ref{sec:gd_convergence} and present the full proof in Appendix \ref{app:gd_convergence}.
\begin{theorem}[GD]\label{thm:gd_convergence}
Assume that $\norm{\bw_0} < 1$ and \checked{$0 < \eta \leq 1 / 5$}. Then the full-batch GD iterates $\bw_t$ converge to the point $\bw_\GD$ as $t \to \infty$ where
\[
\bw_\GD := \frac{\Pi_m(\bw_0)}{\norm{\Pi_m(\bw_0)}},
\]
and the projection $\Pi_m$ is defined in Eq.\ (\ref{eq:proj_pos_set}).  
\end{theorem}
Our result states that full-batch GD converges to the point obtained by taking the initialization $\bw_0$, projecting it onto the span of the data, and then rescaling it to have norm one. Note that this implies that GD is invariant to orthogonal transformations of the data which preserve the span of the dataset.  \checked{In Section~\ref{sec:loss_landscape}, we will show that $\bw_\GD$ is a global minimum of the objective function.}

We will now give the convergence behavior of mini-batch SGD. First define the set of initially positive datapoint directions 
\begin{equation}\label{eq:pos_set}
    \cS^+ := \{i \in [m] : \sip{\bw_0}{\ba_i} > 0\}.
\end{equation}
The following theorem provides a limiting characterization for mini-batch SGD. A sketch of the proof is given in Section \ref{sec:sgd_convergence} and the full proof is given in Appendix \ref{app:sgd_convergence}.
\begin{theorem}[SGD]\label{thm:sgd_convergence}
    Assume $\norm{\bw_0} < 1$ and \checked{$0<\eta \leq 1/5$.} Then the mini-batch SGD iterates $\bw_t$ converge to an element of the set $\{\ba_i: i \in \cS^+\} \cup \{-\ba_j: j \in [m] \setminus \cS^+ \}$  \checked{a.s. over the randomness of SGD minibatch sampling. }
\end{theorem}
\checked{Note that our theorem applies for a deterministic choice of initialization $\bw_0$ and dataset $\cD$ and holds almost surely over the sampling of the mini-batches (recall that at each step of SGD, a batch is sampled uniformly at random from the distinct batches of size $b<m$).} 
Our result does not give the precise probability distribution over the $m$ possible limit points, however we are still able to infer that almost surely $\bw_t$ eventually \checked{converges to a single point and} perfectly aligns itself with some element of the dataset.  \checked{Convergence to a point with a constant step-size is possible due to the fact that all of the pointwise gradients vanish at the SGD limit points.} \checked{Like $\bw_\GD$, the solutions found by SGD are also global minima of the objective function; we shall show this in Section~\ref{sec:loss_landscape}}.

For random initialization $\bw_0 \sim \cN(0, (\sigma_{\init}^2 / n) \cdot \id_n)$, Theorem \ref{thm:gd_convergence} implies that the GD solution $\bw_{\GD}$ is a random unit vector in $\Span(\ba_i : i \in [m])$. If $m$ is large then the iterates will hardly correlate with any particular data direction, but will by highly correlated with the initialization $\bw_0$. This is in contrast to the SGD solution $\bw_{\SGD}$ which by Theorem \ref{thm:sgd_convergence} is perfectly correlated with some datapoint and nearly orthogonal to the intialization. To make this quantitative, let us define for $\bw \in \R^n$ its cosine similarity with $\bx \in \R^n$ which we denote as $\cossim(\bw, \bx)$ and its maximum cosine similarity with the dataset $\cD = \{\ba_1, \ldots, \ba_m\}$ which we denote as $\cossim(\bw, \cD)$, as follows
\begin{equation}\label{eq:maxcorr}
    \cossim(\bw, \bx) := \qty|\left \langle \frac{\bw}{\norm{\bw}}, \frac{\bx}{\norm{\bx}} \right \rangle|,~~\cossim(\bw, \cD) := \max_{\bx \in \cD} \cossim(\bw, \bx).
\end{equation}
We have the following corollary which gives the limiting cosine similarities with the dataset and the initialization for GD and SGD with random initialization.
\begin{corollary}\label{cor:low_correlation}
Assume that $\bw_0 \sim \mathcal{N}(0, (\sigma_{\init}^2 / n) \cdot \id_n)$ where $\sigma_\init < 1$ and $m = \Theta(n)$. If $\eta \leq 1 / 5$, then with probability at least $1 - O(n^{-1})$, the GD iterates $\bw_t^{\GD}$ satisfy
\[
\lim\limits_{t \to \infty} \cossim(\bw_t^{\GD}, \cD) = O\qty(\sqrt{\frac{\log n}{n}}),~~ \lim\limits_{t \to \infty}\cossim(\bw_t^{\GD}, \bw_0) = \Theta(1),
\]
and the iterates of minibatch SGD satisfy
\[
\lim\limits_{t \to \infty} \cossim(\bw_t^{\SGD}, \cD) = 1,~~ \lim\limits_{t \to \infty}\cossim(\bw_t^{\SGD}, \bw_0) = O\qty(\sqrt{\frac{\log n}{n}}),
\]
where $\cossim(\bw_t, \cD)$ is defined in Eq.\ (\ref{eq:maxcorr}). 
\end{corollary}
\checked{Together Theorem \ref{thm:gd_convergence}, Theorem 2, and Corollary \ref{cor:low_correlation} illustrate that SGD finds solutions which are significantly different to its random initialization, while the GD solution is quite similar to its random initialization (as measured by cosine similarity). Furthermore, the GD solution is invariant to orthogonal transformations which preserve the span of the data, which roughly means that the solution only depends on the data through the linear subspace spanned by the data.  All together, these observations support the view that SGD exhibits a stronger form of data-dependent feature learning than GD in this setting. }
\footnote{\checked{We are not aware of an agreed-upon definition of feature learning but a common view in the deep learning literature is that more feature learning occurs if the solution found is far from its initialization and incorporates more data-dependent information~\citep{chizat2018note,woodworth2020kernel}.}}

One may wonder if the convergence behavior of minibatch SGD \checked{in our setting} is primarily driven by the stochasticity of the mini-batch selection. In the next theorem, we suggest this may not be the case and that the more important property is the batch-size. Namely, we show that a totally deterministic cyclic selection of mini-batch indices still leads to convergence to a single datapoint when the iterates are initialized from a certain non-trivial region with positive Lebesgue measure.

\begin{theorem}[CSGD]\label{thm:csgd_convergence}
Let $m = n = 2$ and $\cD = \{\ba_0, \ba_1\}$. Assume that $\ip{\bw_0}{\ba_0} \geq \ip{\bw_0}{\ba_1}$ and $\ip{\bw_0}{\ba_1} > 0$. Furthermore, assume that $\norm{\bw_0} < 1$, and \checked{$0 <\eta \leq 1/4$.} Then the CSGD iterates $\bw_t$ converge to $\ba_0$ as $t \to \infty$.
\end{theorem}
We provide the proof of this result in Appendix \ref{app:csgd_convergence}. Unlike our mini-batch SGD result, here we are able to explicitly determine the convergence point and give support to the intuition that the updates are biased towards converging to the $\ba_i$ the iterate is currently maximally correlated with. Compared to the mini-batch SGD result, however, this result is limited by the fact that we restrict to the two-dimensional setting $m = n = 2$ and impose the initialization condition $\ip{\bw_0}{\ba_0} \geq \ip{\bw_0}{\ba_1}$.  We believe the result should hold more broadly for arbitrary $m, n$ and $\bw_0$ such that $\norm{\bw_0} < 1$ and we have empirically verified this behavior in simulations. However, we do not pursue this direction further in this work as SGD appears to have the same qualitative behavior, SGD is a more common algorithm, and our proof techniques for SGD are more amenable to analysing more general settings.
\subsection{Full-batch GD Proof Sketch}\label{sec:gd_convergence}
In this section we give a proof sketch for Theorem \ref{thm:gd_convergence}. Since the $\ba_i$ are orthonormal we can analyse the evolution of $\bw_t$ in terms of the coordinates $c_t(i) = \ip{\bw_t}{\ba_i}$. The detailed proof is given in Appendix \ref{app:gd_convergence}.   One can write the full-batch gradient descent updates of each coordinate $c_t(i)$ as follows,
\begin{align*}
    c_{t+1}(i) &= c_t(i)(1 +  \eta (2 - 2 \Phi_t - \Psi_t)), && i \in [m]\\
    c_{t+1}(j) &= c_t(j)(1 - \eta \Phi_t), && j \in [n] \setminus [m]
\end{align*}
where we define the quantities
\begin{equation}\label{eq:phi_psi_def}
    \Phi_t := \sum_{i \in [m]} c_t(i)^2,~~\Psi_t := \sum_{j \in [n]\setminus [m]} c_t(j)^2.
\end{equation}
It is easy to see that for full-batch GD, the ratio of the coordinate updates $c_{t+1}(i)/c_t(i)$ is the same for each coordinate $i$.  Using this, we can derive 
the following key invariant,
\begin{equation}\label{eq:gd_invariant}
    c_t(i) = c_0(i)\sqrt{\frac{\Phi_t}{\Phi_0}},~~\text{for all } i \in [m] \text{ and all  } t \in \{0, 1, \ldots\}.
\end{equation}
Thus, in order to understand the dynamics of $c_t(i)$ for $i\in [m]$, it suffices to understand the dynamics of $\Phi_t$.  Similarly, by understanding the dynamics of $\Psi_t$ we can characterize the dynamics of $c_t(j)$ for $j\in [n]\setminus [m]$.  It turns out that $(\Phi_{t+1}, \Psi_{t+1})\in \R^2$ can be written solely in terms of $(\Phi_t, \Psi_t)\in \R^2$, so we instead directly analyze the evolution of this two-dimensional system. 
Under the conditions $\norm{\bw_0} < 1$ and step-size $\eta \leq 1/5$, we can establish the following boundedness property for all $t$,
\begin{equation}\label{eq:gd_norm_bound}
    \Phi_t + (5/8) \Psi_t < 1,~\text{for all } t \in \{0, 1, \ldots\}.
\end{equation}
Using Eq.\ (\ref{eq:gd_norm_bound}), it is not hard to show $\Phi_t \to 1$ and $\Psi_t \to 0$ as $t \to \infty$. Now the result follows since $c_t(i) \to c_0 / \sqrt{\Phi_0}$ for $i \in [m]$ by Eq.\ (\ref{eq:gd_invariant}) and $c_t(j) \to 0$ for $i \notin [m]$.
\subsection{Properties of Mini-batch (S)GD}\label{sec:minibatch_gd}
We now move on to sketching the proof of mini-batch SGD convergence. We start with some general properties of mini-batch (S)GD which hold for any mini-batch sequence \checked{with batch size $b<m$}. The proofs of the results in this section can be found in Appendices \ref{app:minibatch_gd} and \ref{app:minibatch_sgd}. In particular, the proofs of Proposition \ref{prop:bounded_iterates} and Corollary \ref{cor:bounded_coordinates} can be found in Appendix \ref{app:minibatch_gd} and the proof of Proposition \ref{prop:minibatch_sgd} can be found in Appendix \ref{app:minibatch_sgd}. As before we will analyze the evolution of the coordinates $c_t(i)$. From Eq.\ (\ref{eq:minibatch_gd}) and Eq.\ (\ref{eq:grad_single_loss}) we have,
\begin{align*}
    c_{t+1}(i) &= c_{t}(i)\qty(1 + \eta\qty(2 - \norm{\Pi_{\cB_t}(\bw_t)}^2 - \norm{\bw_t}^2)) && i \in \cB_t,\\
    c_{t+1}(j) &= c_t(j)\qty(1 - \eta \norm{\Pi_{\cB_t}(\bw_t)}^2) && j \not\in \cB_t.
\end{align*}
The next result states that for any mini-batch sequence the iterates are bounded in $\ell_2$-norm. 
\begin{proposition}[Bounded Iterates]\label{prop:bounded_iterates}
Assume that $\norm{\bw_0} < 1$ and \checked{$0<\eta \leq 1/5$.} Then for all $t \geq 0$ and \checked{any batch size $b < m$}, the iterates of mini-batch GD for any mini-batch sequence $(\cB_t)_{t \geq 0}$ satisfy
\begin{equation*}
    \norm{\bw_t}^2 \leq 1 + \eta / 4.
\end{equation*}
\end{proposition}
Note that although this bound is weaker than the corresponding one in Eq.\ (\ref{eq:gd_norm_bound}) for full-batch GD, it still provides several useful consequences. One can show that $c_{t+1}(\ell) / c_t(\ell) > 0$ for all $\ell \in [n]$. Thus the coordinates never change sign. Moreover for $j \not\in \cB_t$, $c_{t+1}(j) / c_t(j) < 1$,  hence the magnitude of coordinates not present in the batch decreases each iteration, that is $|c_{t+1}(j)| < |c_t(j)|$. In particular, $\Psi_t$ as defined in Eq.\ (\ref{eq:phi_psi_def}) is decreasing with $t$. Lastly, one can show that the coordinate magnitudes $|c_t(\ell)| < 1$ which is sharper than the $\ell_2$-norm bound from Proposition \ref{prop:bounded_iterates}. We summarize these conclusions in the following corollary.
\begin{corollary}\label{cor:bounded_coordinates}
     Under the conditions of Proposition \ref{prop:bounded_iterates}, for all times $t$ we have $|c_t(i)| < 1$ and $\sign(c_t(i)) = \sign(c_0(i))$ for all $i \in [n]$. Furthermore, $\Psi_t$ is monotonically decreasing.  
\end{corollary}    
So far from Corollary \ref{cor:bounded_coordinates} we know that for any mini-batch sequence that $\Psi_t$ is decreasing and from \ref{prop:bounded_iterates} that $\norm{\bw_t}$ is bounded. The next result shows that under the additional assumption that the mini-batch sequence is chosen at random, we can say more.
\begin{proposition}\label{prop:minibatch_sgd}
For mini-batch SGD with any batch size \checked{$b<m$} the following hold,
\begin{enumerate}
    \item As $t \to \infty$, almost surely $\Psi_t \to 0$,
    \item Almost surely $\liminf_{t \to \infty} \norm{\bw_t} \geq 1$.
\end{enumerate}
\end{proposition}
\checked{Since $\Psi_t$} \checked{as defined in Eq.\ (\ref{eq:phi_psi_def}) is orthogonal projection onto the complement of the data subspace,} the above essentially states that eventually the iterates lie in the subspace spanned by the data and have norm at least one. We will now start to more specifically describe our proof strategy for SGD convergence. 
The proof relies on connecting the convergence of SGD with a particular stochastic process which we describe in the next section.

\subsection{SGD and Random Walks}\label{sec:sgd_random_walk}
The connection between SGD and random walks\footnote{We use the term random walk more generally than to just mean a stochastic process arising from a sequence of partial sums of i.i.d random variables.} arises since, roughly speaking, at each step the neuron $\bw_t$ ``rotates" in the direction of $\ba_i$ in the mini-batch with which it has the highest correlation. We wish to show that asymptotically $\bw_t$ becomes completely aligned with one point $\ba_i$ and completely unaligned with all other datapoints. We can track this relative alignment by analysing a certain one-dimensional random walk which is a function of the iterates.
Let $i_t^\star = \argmax_{i \in [m]} \abs{c_t(i)}$ be the direction with highest alignment and $\cJ_t = [m] \setminus \{i_t^\star\}$ be the set of remaining directions. The random walk we analyse is the log-ratio quantity $\{R_t\}_{t=0}^\infty$ where
\begin{equation}\label{eq:log-ratio}
    R_t := \log\qty(\frac{|c_t(i_t^\star)|}{\sum\limits_{\ell \in \cJ_t} |c_t(\ell)|}).
\end{equation}
Our goal is to show that $R_t \to \infty$ almost surely, as this will imply that the neuron is completely aligned with some direction in the limit as $t\to \infty$. Conveniently, we only need to consider a single ratio involving the most aligned direction $i_t^\star$ since we are not concerned with which particular $\ba_i$ the iterates converge to.
In the next section, we describe tools for analysing stochastic processes of this type.

\subsection{Non-homogeneous Random Walks}\label{sec:random_walk}
We now present a result from the theory of non-homogeneous random walks that is used in our proof of Theorem \ref{thm:sgd_convergence}. We remark that this theory is much broader in scope than what we present and has been used before to analyse other stochastic systems such as urn processes, birth-and-death chains, etc. However, to the best of our knowledge we present the first application of this theory to the analysis of SGD. We believe that such techniques should be broadly useful in analysing the behavior of SGD for other problems.

We now introduce the notation and assumptions. Define $\Z^+ := \{0, 1, \ldots\}$ and let $X = (X_t)_{t \in \Z^+}$ be a discrete time stochastic process adapted to a filtration $(\cF_t)_{t\in \Z^+}$ and taking values in the half-line $[b_0, \infty)$ for some $b_0 \in \R$. Let us state the basic assumptions. 
\begin{enumerate}[label=(A\arabic*)]
    \item\label{enum:irreducibility} For any $y \in (b_0, \infty)$ there exists a function $v : \Z^+ \to \Z^+$ and $\eps > 0$ such that
    \[
        \inf_{t \in \Z^+} \Pr[X_{t+v(t)} > y \mid \cF_t] > \eps,~ a.s.
    \]
    \item\label{enum:moment_condition} For some $K < \infty$ 
    \[
    \sup_{t \in \Z^+} |X_{t+1} - X_t| \leq K,~ a.s.
    \]
\end{enumerate}
The first condition is a type of fairly weak irreducibility condition which states that at any time there is at least some positive probability to exceed a value $y$ after some number of steps and implies in particular that $\limsup_{t \to \infty} X_t = \infty$ a.s. The second condition states the process has bounded increments. Under these conditions we have the following result which is a special case of the more general Theorem 2.2 from \cite{menshikov2010rate} and can be used to show the process $X$ is transient.
\begin{proposition}[\checked{\citet{menshikov2010rate}}]\label{prop:transience}
Let $X = (X_t)_{t \in \Z^+}$ be a stochastic process adapted to the filtration $(\cF_t)_{t \in \Z^+}$ on the half-line $[b_0, \infty)$ for some $b_0 \in \R$. Assume that Assumptions \ref{enum:irreducibility} and \ref{enum:moment_condition} hold for $X$. If there exists a function $\lmu_1 : [b_0, \infty) \to \R$ such that for all $t \in \Z^+$
\[
\lmu_1(X_t) \leq \E[X_{t+1} - X_t \mid \cF_t],~a.s.,
\]
and $\liminf_{x \to \infty} \lmu_1(x) > 0$, then $X$ is transient, that is $X_t \to \infty$ a.s. as $t \to \infty$.
\end{proposition}
\subsection{SGD Proof Sketch}\label{sec:sgd_convergence}
Using the previous results we can now prove Theorem \ref{thm:sgd_convergence}.  
Complete proofs 
for this section can be found in Appendix \ref{app:sgd_convergence}. In the following proposition we establish the transience of the process $R = (R_t)_{t \in \Z^+}$ by verifying that it obeys the conditions of \checked{Proposition~\ref{prop:transience}}. Note that it makes sense to apply this result since 
\begin{equation*}
    R_t = \log\qty(\frac{|c_t(i_t^\star)|}{\sum\limits_{\ell \in \cJ_t} |c_t(\ell)|}) \geq \log\qty(\frac{|c_t(i_t^\star)|}{|\cJ_t||c_t(i_t^\star)|}) = -\log(|\cJ_t|) = -\log(m-1),
\end{equation*}
hence $R$ is a stochastic process on $[b_0, \infty)$ where $b_0 = -\log(m-1)$.
\begin{proposition}[Transience of $R$]\label{prop:log_ratio_transience}
 The process $(R_t)_{t \in \Z^+}$ defined in Eq.\ (\ref{eq:log-ratio}) arising from SGD is transient, i.e., $R_t \to \infty$ a.s. as $t \to \infty$.
\end{proposition}
Let us first try to gain some intuition about this result. Note that for full-batch GD we actually have that $R_{t} = R_0$ for all $t$, hence this result is not true for full-batch GD. This is because for full-batch GD $c_{t+1}(i)/c_t(i) = 1 + \eta(2 - 2\Phi_t - \Psi_t)$ for all $i \in [m]$, hence $R_{t+1} = R_t$. 
Now let us try to see why the process is transient for mini-batch SGD. 
\paragraph{Sketch of Proposition \ref{prop:log_ratio_transience}}
To invoke \checked{Proposition~\ref{prop:transience}} we will need to analyze the increment of the process $R_{t+1} - R_t$. More specifically, we will need to show that the conditional expected increment is lower bounded by a positive constant, in addition to verifying Assumptions \ref{enum:irreducibility} and \ref{enum:moment_condition}. Let us first observe that the increment $R_{t+1} - R_t \geq \Delta_t$ where
\begin{equation}\label{eq:delta_lower_bound}
    \Delta_t := \log\qty(\frac{|c_{t+1}(i_t^\star)|}{\sum_{\ell \in \cJ_t} |c_{t+1}(\ell)|}) - \log\qty(\frac{|c_{t}(i_t^\star)|}{\sum_{\ell \in \cJ_t} |c_{t}(\ell)|}).
\end{equation}
This follows since $|c_{t+1}(i_{t+1}^\star)| \geq |c_{t+1}(i_t^\star)|$ by definition and
\begin{equation*}
    \sum_{\ell \in \cJ_{t+1}} |c_{t+1}(\ell)| = \sum_{\ell \in [m]} |c_{t+1}(\ell)| - |c_{t+1}(i_{t+1}^\star)| \leq \sum_{\ell \in [m]} |c_{t+1}(\ell)| - |c_{t+1}(i_t^\star)| = \sum_{\ell \in \cJ_t} |c_{t+1}(\ell)|,
\end{equation*}
hence the first term on the right of Eq.\ (\ref{eq:delta_lower_bound}) is less than $R_{t+1}$ and the second term is just $R_t$. Thus to lower bound the increment, it will suffice to lower bound the more tractable quantity $\Delta_t$. Intuitively, we should expect that $\Delta_t$ will be positive when $i_t^\star \in \cB_t$ and negative otherwise. By considering these two cases we will show that in expectation $\Delta_t$ is positive. Importantly however, note that from the statement of \checked{Proposition~\ref{prop:transience}} that we only need to establish this lower bound \textit{asymptotically} for large $R_t$ and large times $t$. For the purposes of the proof sketch we will informally use the notation $A \gtrapprox B$ to denote that the inequality is true up to an error which vanishes as $R_t \to \infty$ and $t \to \infty$, with $A \approx B$ taken to mean $A \gtrapprox B$ and $A \lessapprox B$. In particular, we have the following asymptotic statement
\begin{equation}\label{eq:asymp}
    \norm{\bw_t}^2 \approx c_t(i_t^\star)^2 \approx 1.
\end{equation}
To see why this holds note that since $|c_t(\ell)| < 1$ by Corollary \ref{cor:bounded_coordinates},
\begin{equation*}
    \norm{\bw_t}^2 - c_t(i_t^\star)^2 - \Psi_t = \sum\limits_{\ell \in \cJ_t} c_t(\ell)^2  \leq \sum\limits_{\ell \in \cJ_t} |c_t(\ell)| = |c_t(i_t^\star)| \exp(-R_t) \leq \exp(-R_t).
\end{equation*}
By Proposition \ref{prop:minibatch_sgd}, $\Psi_t \to 0$, hence from the above $\norm{\bw_t}^2 \approx c_t(i_t^\star)^2$. Also, $\liminf \norm{\bw_t}^2 \geq 1$ hence $\norm{\bw_t}^2  \gtrapprox 1$. Since $c_t(i_t^\star)^2 \leq 1$ this yields Eq.\ (\ref{eq:asymp}).

Now we move on to lower bounding $\E(\Delta_t \mid \cF_t)$ by a quantity which becomes a time-independent positive constant as $R_t \to \infty$ and $t \to \infty$. For simplicity, in this sketch let us consider the setting where $b = 1$ and $c_0(\ell) > 0$ for all $\ell \in [n]$. By Corollary \ref{cor:bounded_coordinates}, $c_t(\ell) \in (0, 1)$ for all $t$. Let $i_t \in [m]$ be the selected mini-batch index. With probability $1/m$, we have $i_t = i_t^\star$. In this case one can calculate that
\begin{equation}\label{eq:delta_positive}
    \Delta_t =  \log\qty(\frac{1 + \eta(2 - c_t(i_t^\star)^2 - \norm{\bw_t}^2)}{1 - \eta c_t(i_t^\star)^2}) \approx \log\qty(\frac{1}{1 - \eta})
\end{equation}
where we used that $\norm{\bw_t}^2 \approx c_t(i_t^\star)^2 \approx 1$. On the other hand, if $i_t = i$ for some $i \neq i_t^\star$ then 
\begin{equation*}
    -\Delta_t =  \log\qty(\frac{1}{1 - \eta c_t(i_t)^2} \frac{\sum_{\ell \in \cJ_t} c_{t+1}(\ell) }{\sum_{\ell \in \cJ_t} c_{t}(\ell)})
\end{equation*}
where we can expand the term
\begin{align*}
    \frac{\sum_{\ell \in \cJ_t} c_{t+1}(\ell)}{\sum_{\ell \in \cJ_t} c_{t}(\ell)} &= \frac{\eta c_t(i_t)(1 + \eta(2 - c_t(i_t)^2 - \norm{\bw_t}^2)) + (1 - \eta c_t(i_t)^2) \sum_{\ell \in \cJ_t \setminus \{i_t\}} c_t(\ell)}{\sum_{\ell \in \cJ_t} c_t(\ell)} \\
    &= \frac{c_t(i_t)(2 - \norm{\bw_t}^2) + (1 - \eta c_t(i_t)^2) \sum_{\ell \in \cJ_t} c_t(\ell)}{\sum_{\ell \in \cJ_t} c_t(\ell)}\\
    &\approx \frac{\eta c_t(i_t)}{\sum_{\ell \in \cJ_t} c_t(\ell)} + (1 - \eta c_t(i_t)^2).
\end{align*}
Combining with the earlier expression and using that $c_t(i_t)^2 \approx 0$ by Eq.\ (\ref{eq:asymp}) gives 
\begin{align*}
    -\Delta_t &\approx  \log\qty(1 + \frac{1}{1 - \eta c_t(i_t)^2} \frac{\eta c_t(i_t)}{\sum_{\ell \in \cJ_t} c_t(\ell)})\\
    &\approx \log\qty(1 + \frac{\eta c_t(i_t)}{\sum_{\ell \in \cJ_t} c_t(\ell)})\\
    &\leq \eta \cdot \frac{c_t(i_t)}{\sum_{\ell \in \cJ_t} c_t(\ell)}. 
\end{align*}
Therefore combining with Eq.\ (\ref{eq:delta_positive}) we have that conditioned on $\cF_t$
\begin{align*}
    \E \Delta_t &\gtrapprox \Pr(i_t = i_t^\star) \log\qty(\frac{1}{1 - \eta}) - \sum_{i \in \cJ_t} \Pr(i_t = i) \cdot \eta \cdot \frac{c_t(i)}{\sum_{\ell \in \cJ_t} c_t(\ell)}\\
    &= \frac{1}{m}\qty[\log\qty(\frac{1}{1 - \eta}) - \eta] \geq \frac{\eta^2}{2m},
\end{align*}
where the last inequality uses  $\log(1 / (1-x)) \geq x(1+x/2)$ for $x \in (0, 1)$. This accomplishes our goal of asymptotically bounded the expected conditional increment. It is not much harder to verify Assumptions \ref{enum:irreducibility} and \ref{enum:moment_condition}. To see that \ref{enum:irreducibility} holds, note that if $i_t = i_t^\star$, then from Eq.\ (\ref{eq:delta_positive}), using that $\norm{\bw_t}^2 \leq 1 + \eta/4$ from Proposition \ref{prop:bounded_iterates} yields
\begin{align*}
    \Delta_t &= \log\qty(\frac{1 + \eta(2 - c_t(i_t^\star)^2 - \norm{\bw_t}^2)}{1 - \eta c_t(i_t^\star)^2})\\
    &= \log(1 + \eta \frac{2 - \norm{\bw_t}^2}{1 - \eta c_t(i_t^\star)^2})\\ 
    &\geq \log(1 + \eta(1 - \eta/4)) > 0,
\end{align*}
where the last inequality holds since $\eta(1 - \eta / 4) > 0$. Thus on this event we have lower bounded the increment by a time-independent constant. Furthermore the probability of this event is $1/m$ which is also time-independent. Therefore we can see that Assumption (A1) will be satisfied by considering the event that $i_t = i_t^\star$ a sufficiently large (but time-independent) number of times in a row. Verifying Assumption \ref{enum:moment_condition} is not very difficult and should be plausible given that $|c_t(\ell)| < 1$ and $\norm{\bw_t}^2 \leq 1 + \eta / 4$.
 \paragraph{Sketch of Theorem \ref{thm:sgd_convergence}}
 Now taking Proposition \ref{prop:log_ratio_transience} to be true, the rest of the proof follows quite easily. By Corollary \ref{cor:bounded_coordinates},
 \begin{align*}
        1 \geq c_t(i_t^\star)^2 &= \norm{\bw_t}^2 - \sum\limits_{\ell \in \cJ_t} c_t(\ell)^2 - \Psi_t\\
        &\geq \norm{\bw_t}^2 - \exp(-R_t) - \Psi_t.
    \end{align*}
Now by Propositions \ref{prop:minibatch_sgd} and \ref{prop:log_ratio_transience}, we have $\liminf \norm{\bw_t} \geq 1$, $\Psi_t \to 0$, and $R_t \to \infty$ as $t \to \infty$, so we see that $|c_t(i_t^\star)| \to 1$.  One can then show that $i_t^\star$ must eventually become constant since the gradient norm goes to zero when $\bw_t$ approaches any $\ba_i$. Therefore there exists some $i^\star \in [m]$ such that $i_t^\star = i^\star$ eventually. By Corollary \ref{cor:bounded_coordinates} we have that $\sign(c_t(i^\star)) = \sign(c_0(i^\star))$, hence $\bw_t \to \sign(c_0(i^\star)) \cdot \ba_i$ as we wished to show.

\section{Theoretical Results for ReLU Activation}\label{sec:theory_relu_act}
In this section, we will consider the case when the activation function of the autoencoder in Eq.~\eqref{eq:autoencoder_definition} is the ReLU $\phi(t) = \max(t, 0)$.\footnote{\checked{Note that the ReLU is non-differentiable at the origin, although a sub-gradient exists for $\phi(t)$ at every $t\in \R$.  The only issue that could arise is if the weights are exactly orthogonal to one of the $\ba_i$, but with a Gaussian random initialization this does not occur almost surely.}}  Our results will rely upon \checked{an equivalence} between the dynamics of SGD/GD for ReLU autoencoders with the dynamics of SGD/GD for linear autoencoders.  To this end, let us introduce some preliminary notation.  Let us denote the output and losses for the ReLU autoencoder and linear autoencoder as,
\begin{align*}
    f(\bx; \bw) &= \bw \max(\sip{\bw}{\bx}, 0)~~&\ell(\bw; \bx) = \frac{1}{2}\norm{\bx - f(\bx; \bw)}^2 && \text{(ReLU autoencoder)},\\
    \wt{f}(\bx; \bw) &= \bw \sip{\bw}{\bx}~~&\wt{\ell}(\bw; \bx) = \frac{1}{2}\norm{\bx - \wt{f}(\bx; \bw)}^2 && \text{(Linear autoencoder)}.
\end{align*}
Let 
\[ \cS_t^+ = \{i \in [m] : \sip{\ba_i}{\bw_t} > 0\},\qquad \wt{\cB}_t := \cB_t \cap \cS_t^+.\]
The set $\cS_t^+$ consists of datapoints which the neuron is positively correlated with at time $t$, and $\wt{\cB}_t$ is the subset of the batches selected at time $t$ which are also in $\cS_t^+$.   Our key observation is that the minibatch GD update for the ReLU autoencoder with minibatch $\cB_t$ is equivalent to the minibatch GD update if the activation was linear and the minibatch was instead the set $\wt{\cB}_t = \cB_t \cap \cS_t^+$.  That is,
the update in Eq.\ (\ref{eq:minibatch_gd}) satisfies 
\begin{equation}\label{eq:relu_to_linear}
    \bw_t - \eta \sum\limits_{i \in \cB_t} \grad_{\bw}~ \ell(\bw_t; \ba_i) = \bw_t - \eta \sum\limits_{i \in \wt{\cB}_t} \grad_{\bw}~ \wt{\ell}(\bw_t; \ba_i),
\end{equation}
simply since for $i \in [m]$,
\[
\grad_{\bw}~ \ell(\bw; \ba_i)
= 
\begin{cases}
    \grad_{\bw}~ \wt{\ell}(\bw; \ba_i) & \text{if } i \in \cS_t^+,\\
    \bzero & \text{otherwise}.
\end{cases}
\]

By Corollary \ref{cor:bounded_coordinates} in Section \ref{sec:minibatch_gd}, if we define $\cS^+ := \cS^+_0$ then we know that if $\norm{\bw_0} < 1$ and $\eta \leq 1/5$, then for any minibatch GD algorithm $\cS_t^+ = \cS^+$ for all $t$ and so $\wt{\cB}_t = \cB_t \cap \cS^+$.

For full-batch GD, $\wt{\cB}_t = [m] \cap \cS^+ = \cS^+$, hence full-batch GD on a ReLU autoencoder with initialization $\bw_0$ is equivalent to running full-batch GD on a linear autoencoder with initialization $\bw_0$ but with the dataset $\cS^+$ instead of $[m]$. Thus from Theorem \ref{thm:gd_convergence} it is easy to see that we have the following theorem for GD convergence for a ReLU autoencoder.

\begin{theorem}[GD-ReLU]\label{thm:gd_relu}
Assume that $\norm{\bw_0} < 1$ and $\eta \leq 1/5$. Define the set $\cS^+$ as in Eq.\ (\ref{eq:pos_set}). Then the full-batch GD iterates $\bw_t$ converge to the point $\bw_{\GD}$ as $t \to \infty$ where
\begin{equation*}
    \bw_{\GD} := \frac{\Pi_{\cS^+}(\bw_0)}{\norm{\Pi_{\cS^+}(\bw_0)}}.
\end{equation*}
\end{theorem}
For the case of mini-batch SGD, note that $|\wt{\cB}_t| \leq |\cB_t| < m$. If $|\wt{\cB}_t| = 0$, then nothing happens that iteration, and we can just focus on the subsequence where $|\wt{\cB}_t| > 0$. By Eq.\ (\ref{eq:relu_to_linear}) we see that
minibatch SGD on the ReLU autoencoder is equivalent to a minibatch GD algorithm on a linear autoencoder with dataset $\cS^+$ where the minibatch $\wt{\cB}_t \subseteq \cS^+$ has a random batch-size that can vary with time.  More specifically, we can view the process of selecting $\wt{\cB}_t$ as first randomly choosing the effective batch size $\tilde{b}_t := |\wt{\cB}_t| \in \{1, \ldots, b\}$, and then conditioned on this choice $\wt{\cB}_t$ is chosen uniformly from the set of subsets of $\cS^+$ of size $\tilde{b}_t$, that is, $\wt{\cB}_t$ is a batch size $\tilde{b}_t$ minibatch SGD selection from $\cS^+$. 

With the above in mind, we can essentially transfer the minibatch SGD proof for the linear case to the ReLU setting by viewing $\cS^+$ as the effective dataset. Accordingly, one can show that the stochastic process 
\begin{equation*}
    \wt{R}_t := \log\qty(\frac{c_t(i_t^\star)}{\sum\limits_{j \in \cS^+ \setminus \{i_t^\star\} } c_t(j)}),~~i_t^\star := \argmax_{i \in \cS^+} c_t(i),
\end{equation*}
which arises from replacing $[m]$ with $\cS^+$ in the definition of $R_t$ in Eq.\ (\ref{eq:log_ratio_def}) is transient. Note that the proof of the transience of the stochastic process $R$ in Proposition \ref{prop:log_ratio_transience} relied only on showing that the properties of the increment $R_{t+1} - R_t$ required by \checked{Proposition~\ref{prop:transience}} hold, which was done for any batch size $b < m$ (see Appendix \ref{app:log_ratio_transience}). Thus these properties hold conditionally on $\tilde{b}_t$, from which it is not hard to see that they extend to hold unconditionally as well, hence the process $\wt{R}_t$ is indeed transient. From there, following the same logic as in the rest of the proof of Theorem \ref{thm:sgd_convergence} (see Appendix \ref{app:sgd_thm_proof}), it is easy to see that the following holds for ReLU autoencoder.
\begin{theorem}[SGD-ReLU]\label{thm:sgd_relu}
    Assume that $\norm{\bw_0} < 1$ and \checked{$0<\eta \leq 1/5$.}  Then the mini-batch SGD iterates $\bw_t$ converge to some element of the set $\{\ba_i : i \in \cS^+\}$ almost surely.
\end{theorem}

\subsection{Loss Landscape}\label{sec:loss_landscape}
\begin{figure}[ht]
\vskip 0.2in
\begin{center}
\centerline{\includegraphics[width=0.6\columnwidth]{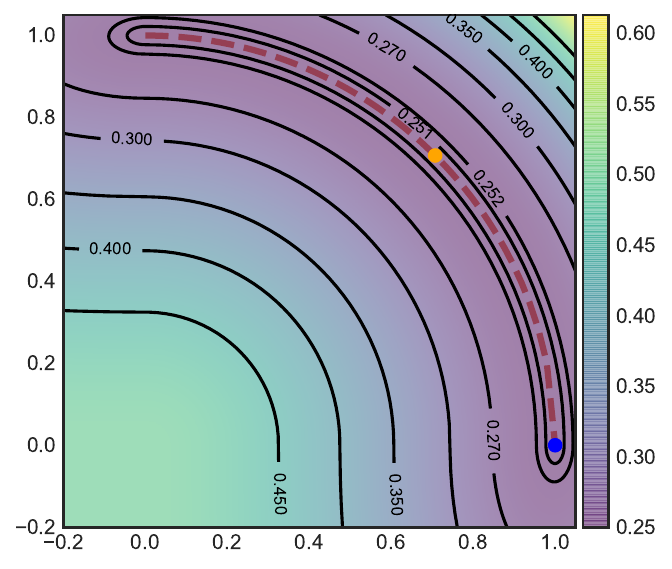}}
\caption{Contour plot of the loss objective $\cL(\bw)$ where $m=n=2$ and $\ba_1=(1,0)^\top$ and $\ba_2=(0,1)^\top$. The red dotted line shows the set of global minima $\cM$ given in Theorem \ref{thm:global_minima}. The blue circle shows one possible solution that can be obtained with SGD ($\bw_{\SGD}=\ba_1$), whereas the orange circle shows a solution that can be obtained with GD but not with SGD ($\bw_{\GD}=1/\sqrt{2} \cdot\ba_1 + 1/\sqrt{2} \cdot\ba_2$).}
\label{fig:contour_plot}
\end{center}
\vskip -0.2in
\end{figure}

In this section we will study properties of the loss landscape of the ReLU autoencoder at the points which GD and SGD converge to. 
\checked{Our first result characterizes the set of global minima of the loss objective.}
\begin{theorem}[Global Minima]\label{thm:global_minima}
The minimum value of the loss objective $\cL(\bw)$ from Eq.\ (\ref{eq:autoencoder_objective}) is attained on
\[
\cM = \l\{\sum_{i=1}^{m}c_i \ba_i: c_1,\ldots,c_m \geq 0 \text{ and } \sum\limits_{i = 1}^m c_i^2 = 1\r\},
\]
where it achieves the value $(m-1) / (2m)$.
\end{theorem}
A visualization of the loss landscape is given in Figure~\ref{fig:contour_plot} when $m = n =2$. We note that a \checked{similar} argument used to prove Theorem~\ref{thm:global_minima} shows that for the case of the linear autoencoder, the global minima are attained on the set $\widetilde\cM = \{ \summ i m c_i \ba_i : \sum_{i=1}^m c_i^2=1\}$, which is defined just like $\cM$ except the coefficients $c_i$ are not required to be non-negative. 

By the result above and our convergence theorems we can see that both GD and SGD converge to global minima. Indeed, Theorem \ref{thm:gd_relu} shows that full batch gradient descent converges to the following solution
\begin{equation}\label{eq:gd_critical_point}
    \bw_{\GD} = \sum_{i \in S} \frac{\ip{\bw_0}{\ba_i}}{\sqrt{\Phi}} \ba_i, ~~~~\Phi = \sum_{i \in \cS^+} \ip{\bw_0}{\ba_i}^2
\end{equation}
where $\cS^+$ is defined in Eq.\ (\ref{eq:pos_set}). By the above theorem $\bw_{\GD}$ is a global minimum. From Theorem \ref{thm:sgd_relu} we see that SGD converges to
\begin{equation}\label{eq:sgd_critical_point}
    \bw_{\SGD} = \ba_i,~~ \text{ for some } i \in \cS^+.
\end{equation}
Again by Theorem \ref{thm:global_minima} this point is also a global minimum. Thus, these algorithms optimally minimize the loss objective, but achieve qualitatively different solutions. 
For random intializations, SGD learns a ``pure" datapoint whereas GD learns a ``mixture".

 In recent years, much of the literature on deep learning has sought to distinguish the learned solutions of large and small batch GD by the ``sharpness" of the obtained solution. The prevailing intuition is that small batch sizes lead to flatter minima (which is correlated with better generalization). However, we show that for our problem, this is not true for common measures of sharpness which either fail to distinguish the two types of solutions or lead to the conclusion that smaller batches lead to \emph{sharper} minima. \checked{Hence, the claim that smaller batch sizes leads to sharper minima is not true without adding further assumptions or adjusting the definition of sharpness.}

The measures of sharpness we consider come from the eigenspectrum of the Hessian. Namely, if $\bH$ is the Hessian of the loss at a given point, then we consider either the maximal eigenvalue $\norm{\bH}_2$ or the sum of the eigenvalues $\Tr(\bH)$, where larger values indicate sharper points. The following result gives the sharpness at the points $\bw_\GD$ and $\bw_\SGD$.
\begin{theorem}\label{thm:sharpness}
Denote the Hessians\footnote{As explained in Appendix \ref{app:loss_landscape}, due to the non-differentiability of the ReLU function, we consider more general measures based on one-sided derivatives which extend the Hessian based definitions.} of the loss $\cL(\bw)$ at the points $\bw_\GD$  and $\bw_\SGD$ defined in Eq.\ (\ref{eq:gd_critical_point}), (\ref{eq:sgd_critical_point}) as $\bH_{\GD}$ and $\bH_{\SGD}$ respectively. Then,
\begin{align*}
    \norm{\bH_\GD}_2 &= \frac{4}{m},~~ \Tr(\bH_{\GD}) = \frac{2n + 8 - m - |\cS^+|}{2m},\\ 
    \norm{\bH_{\SGD}}_2 &= \frac{4}{m},~~
    \Tr(\bH_{\SGD}) = \frac{2n + 7 - m}{2m}.
\end{align*}
\end{theorem}
From the above we have the following observations. Both $\bH_{\GD}$ and $\bH_{\SGD}$ have the same maximum eigenvalue, hence this measure fails to distinguish $\bw_{\GD}$ and $\bw_{\SGD}$. For large $m$, if $\bw_0$ is initialized by a random Gaussian then with high probability $|\cS^+| \approx m / 2$, hence
\[
    \Tr(\bH_{\GD}) = \frac{2n + 8 - m - |\cS^+|}{2m} \approx \frac{2n + 8 - m - m/2}{2m}
\]
from which it follows
\[
\Tr(\bH_{\GD}) \approx \Tr(\bH_{\SGD}) - \frac{1}{2} < \Tr(\bH_{\SGD}),
\]
hence $\bw_{\SGD}$ is \emph{sharper} than $\bw_{\GD}$ as claimed.

%% file: sections/conclusion.tex
In this work, we investigated the dynamics of SGD and GD for training single neuron autoencoders with ReLU activation following random initialization \checked{on orthogonal data.} We showed that for any choice of batch size, SGD/GD converge to global minima despite nonconvexity.  However, the particular minimum found depends strongly upon the batch size.  In the full-batch deterministic setting, GD converges to a minimum which is 
\checked{dense (i.e., not sparse) with respect to the training data}
and is highly aligned with its initial direction.  
\checked{As such, relatively little feature-learning occurs in the full-batch setting.}
For any batch size strictly smaller than the number of samples, SGD converges to a sparse global minimum 
which is almost orthogonal to its initialization\checked{, hence SGD exhibits stronger feature learning compared to GD}.
Moreover, SGD finds solutions which are sharper than those found by full-batch GD if we measure sharpness by the trace of the Hessian. \checked{We are able to prove that the SGD iterates converge almot surely to a degenerate point mass distribution even with a constant step size by introducing and using tools from the literature on non-homogeneous random walks. We believe that these tools may be more broadly applicable for the analysis of other machine learning algorithms.}   

It is worth emphasizing that although we find that the minima found by SGD can be distinguished from those found by GD by sharpness, the relationship we find (SGD producing sharper minima) is the opposite of that found by prior work~\citep{keskar2016large}.  This suggests that sharpness may not be the most useful way to characterize the effect of batch size in the dynamics of SGD.  On the other hand, we also show that at least for single neuron autoencoders, the effect of batch size can be distinguished through the lens of sparsity and through the lens of feature learning: we find that SGD prefers sparse minima and that stochastic gradients lead to significantly different features than those found at random initialization.

For future work, we are interested in understanding the effect of the batch size on the dynamics of SGD/GD in more complex models and distributions.  A natural question is whether smaller batches produce similar effects in multi-neuron autoencoders trained on more general sparse coding \checked{data} models.

%% file: sections/appendix.tex
In the appendix we provide missing proofs from the main paper and some additional background on non-homogeneous random walk theory. In appendix sections \ref{app:minibatch_gd}, \ref{app:minibatch_sgd}, \ref{app:sgd_convergence}, \ref{app:gd_convergence}, and \ref{app:csgd_convergence} we provide convergence results for the linear autoencoder. In Appendix \ref{app:minibatch_gd} we give general results concerning the iterates of minibatch GD for any minibatch sequence and in Appendix \ref{app:minibatch_sgd} we give similar results under the additional assumption that the minibatch indices are chosen randomly. Then in Appendix \ref{app:minibatch_sgd} we prove the SGD convergence result Theorem \ref{thm:sgd_convergence}, in Appendix \ref{app:gd_convergence} we prove the GD convergence result Theorem \ref{thm:sgd_convergence}, and in Appendix \ref{app:csgd_convergence} we prove the CSGD convergence result Theorem \ref{thm:csgd_convergence}. Finally in Appendix \ref{app:loss_landscape} we provide proofs for results about the loss landscape of the ReLU autoencoder.

\section{General Properties of Minibatch Gradient Descent}\label{app:minibatch_gd}
In this section we will present some general properties of the iterates of minibatch gradient descent on a linear autoencoder (i.e., $\phi(t) = t$ in Eq.\ (\ref{eq:autoencoder_definition})) for an arbitrary sequence of minibatch indices, which will be useful for later sections. First let us establish some notation. For a set $\cS \subseteq [n]$ define the orthogonal projection onto the directions in $\cS$ as
\begin{equation*}
    \Pi_{\cS}(\bx) := \sum\limits_{i \in \cS} \ip{\ba_i}{\bx} \bx,
\end{equation*}
with $\Pi_m := \Pi_{[m]}$. Define the coordinate $c_i(t) := \ip{\bw_t}{\ba_i}$ for $i \in [n]$. We will define
\begin{align}
    \Phi_t &:= \norm{\Pi_m(\bx)}^2 = \sum\limits_{i \in [m]} c_t(i)^2,\label{eq:phi_def}\\
    \Psi_t &:= \norm{\bw_t}^2 - \norm{\Pi_m(\bx)}^2 = \sum\limits_{j \in [n] \setminus [m]} c_t(j)^2.\label{eq:psi_def}
\end{align}
Let us write the minibatch GD update in these coordinates. We shall repeatedly use this formulation in the remaining proofs. The updates are given in the following lemma.
\begin{lemma}[Coordinate Updates]\label{lem:coordinate_updates}
Consider the minibatch GD updates given in Eq.\ (\ref{eq:minibatch_gd}). If we define $c_t(\ell) = \ip{\bw_t}{\ba_\ell}$ for $\ell \in [n]$, then the updates are equivalently given by
 \begin{align*}
    c_{t+1}(i) &= c_t(i)(1 + \eta(2 - u_t - \norm{\bw_t}^2)) && i \in \cB_t,\\ 
    c_{t+1}(j) &= c_t(j)(1 - \eta u_t) && j \in [n] \setminus \cB_t,
\end{align*}   
where we define $u_t := \norm{\Pi_{\cB_t}(\bw_t)}^2$.
\end{lemma}

\begin{proof}
We would like to derive equations defining $c_{t+1}(k) = \sip{\bw_{t+1}}{\ba_k}$ for each $k$.  This will depend upon whether or not $k\in \cB_t$.    Using the definition of the minibatch GD updates in Eq.~\eqref{eq:minibatch_gd}, for any $i,\, k\in [m]$ we have,
 \begin{align*}
     \sip{\nabla \ell(\bw; \ba_i)}{\ba_k} &= \sip{\bw \ba_i \bw^\top + \sip{\bw}{\ba_i} \id_n ( \bw \sip{\bw}{\ba_i} - \ba_i) }{\ba_k} \\
     &= \sip{\ba_i}{\ba_k} \snorm{\bw}^2 \sip{\bw}{\ba_i} - \sip{\ba_i}{\bw} \sip{\ba_i}{\ba_k} + \sip{\bw}{\ba_k} \sip{\bw}{\ba_i}^2 - \sip{\bw}{\ba_i} \sip{\ba_i}{\ba_k} \\ 
     &= \sip{\bw}{\ba_i} \l[ \sip{\ba_i}{\ba_k} \l( \snorm{\bw}^2 -2 \r) + \sip{\bw}{\ba_k} \sip{\bw}{\ba_i} \r].
 \end{align*}
 From here, we see that
 \begin{align*}
     \sip{\bw_{t+1}}{\ba_k} &= \sip{\bw_t}{\ba_k} - \eta \sum_{i\in \cB_t} \sip{\nabla \ell(\bw_t; \ba_i)}{\ba_k} \\ 
     &\overset{(i)}= \sip{\bw_t}{\ba_k} - \eta \Bigg[ \sip{\bw_t}{\ba_k} \l( \snorm{\bw_t}^2 - 2  + \sip{\bw_t}{\ba_k}^2 \r) \ind(k\in \cB_t)+ \sip{\bw_t}{\ba_k}\sum_{i\in \cB_t,\, i\neq k} \sip{\bw_t}{a_i}^2  \Bigg] .
 \end{align*}
Equality $(i)$ uses that the $\ba_\ell$'s are orthogonal.  Thus, if $k\not \in \cB_t$, we see that 
 \begin{align*}
     \sip{\bw_{t+1}}{\ba_k} &=\sip{\bw_t}{\ba_k}\l( 1 - \eta \sum_{i\in \cB_t} \sip{\bw_t}{a_i}^2  \r) = \sip{\bw_t}{\ba_k}\l( 1 -u_t  \r).
 \end{align*}
 On the other hand, if $k\in \cB_t$, we have,
 \begin{align*}
     \sip{\bw_{t+1}}{\ba_k}&= \sip{\bw_t}{\ba_k} - \eta \Bigg[ \sip{\bw_t}{\ba_k} \l( \snorm{\bw_t}^2 - 2  + \sip{\bw_t}{\ba_k}^2 \r)+ \sip{\bw_t}{\ba_k} \sum_{i\in \cB_t,\, i\neq k} \sip{\bw_t}{a_i}^2\Bigg]\\ 
     &= \sip{\bw_t}{\ba_k} \l( 1 - \eta \l( \snorm{\bw_t}^2 + u_t -2\r) \r).
 \end{align*}
Putting these together, we see that the updates of $c_t(i)$ for $i\in [n]$ can be written as follows:
\begin{align*}
    c_{t+1}(i) &= c_t(i)(1 + \eta(2 - u_t - \norm{\bw_t}^2)) && i \in \cB_t,\\ 
    c_{t+1}(j) &= c_t(j)(1 - \eta u_t) && j \in [n] \setminus \cB_t.
\end{align*}
\end{proof}
 
Throughout, we will always be operating under the following assumption on the initialization and step-size of minibatch gradient descent.
\begin{assumption}\label{ass:init}
    Assume that $\norm{\bw_0} < 1$ and $\eta \leq 1/5$.
\end{assumption}
Under this assumption we show that the minibatch GD iterates are bounded above in norm in the following proposition which is a restatement of Proposition \ref{prop:bounded_iterates} from the main paper.
\begin{proposition}[Bounded Iterates]\label{prop:bounded_iterates_app}
    For all $t$ the iterates of Eq.\ (\ref{eq:minibatch_gd}) for any mini-batch sequence $(\cB_t)_{t \geq 0}$ satisfy $\norm{\bw_t}^2 \leq 1 + \eta / 4$.
\end{proposition}
\begin{proof}
We prove this by induction. Clearly the statement holds at $t = 0$ by Assumption \ref{ass:init}. Now assume the statement holds at time $t$. For convenience define $N_t := \norm{\bw_t}^2$. Since the data is orthonormal we have
    \begin{align*}
        N_{t+1} &= \sum\limits_{i \in \cB_t} c_{t+1}(i)^2 + \sum\limits_{j \in [n]\setminus \cB_t} c_{t+1}(j)^2\\
        &= \sum\limits_{i \in \cB_t} c_{t}(i)^2(1 + \eta(2 - u_t - N_t))^2 + \sum\limits_{j \in [n]\setminus \cB_t} c_{t}(j)^2(1 - \eta u_t)^2\\
        &= u_t(1 + \eta(2 - u_t - N_t))^2 + (N_t - u_t)(1 - \eta u_t)^2.
    \end{align*}
    Let us consider the following function 
    \[
        f(N, u) = u(1 + \eta(2 - u - N))^2 + (N - u)(1 - \eta u)^2.
    \]
    Let $N_{\max} := 1 + \eta / 4$. It suffices to show that
    \[
    \max_{N, u \in [0, N_{\max}]} f(N, u) \leq N_{\max}.
    \]
    Note that if $u \geq 0$, then $f(N, u)$ is a convex quadratic in $N$ hence 
    \begin{align*}
        \max_{N, u \in [0, N_{\max}]} f(N, u) &= \max_{u \in [0, N_{\max}]} \max_{N \in [0, N_{\max}]} f(N, u)\\
        &= \max_{u \in [0, N_{\max}]} \{\max(f(0, u), f(N_{\max}, u))\}\\
        &= \max\qty(\max_{u \in [0, N_{\max}]} f(0, u), \max_{u \in [0, N_{\max}]} f(N_{\max}, u)).
    \end{align*}
    Therefore it suffices to show that
    \[
        \max_{u \in [0, N_{\max}]} f(0, u) \leq N_{\max}~~ \text{ and } ~~\max_{u \in [0, N_{\max}]} f(N_{\max}, u) \leq N_{\max}.
    \]
    Plugging in $N=0$ we have
    \[
        \max_{u \in [0, N_{\max}]} f(0, u)
         = \max_{u \in [0, N_{\max}]} 4\eta u(1 + \eta(1 - u)).
    \]
    Since $u \mapsto 4\eta u(1 + \eta(1 - u))$ is increasing for $u \leq (1+\eta) / 2\eta$ and $N_{\max} \leq (1+\eta) / 2\eta$,
    \begin{align*}
        \max_{u \in [0, N_{\max}]} 4\eta u(1 + \eta(1 - u)) &= 4 \eta N_{\max}(1 + \eta(1 - N_{\max}))\\
        &\leq N_{\max}(1 - \eta^2/4) \leq N_{\max}.
    \end{align*}
    We can bound the other term $\max_{u \in [0, N_{\max}]} f(N_{\max}, u)$ as follows
    \begin{align*}
        \max_{u \in [0, N_{\max}]} f(N_{\max}, u) &= \max_{u \in [0, N_{\max}]} u(1 + \eta(2 - u - N_{\max}))^2 + (N_{\max} - u)(1 - \eta u)^2\\
        &= \max_{u \in [0, N_{\max}]} \eta u(2 - N_{\max})(2 - 2 \eta u + \eta(2 - N_{\max})) + N_{\max}(1 - \eta u)^2\\
        &= \max_{u \in [0, N_{\max}]} \eta^2(3 N_{\max}-4)u^2 + \eta u[\eta(2-N_{\max})^2 + 4(1-N_{\max})] + N_{\max}
    \end{align*}
    Observe that the first and second terms in the last line are non-positive since
    \begin{align*}
        &3N_{\max}- 4 = 3 + 3 \eta/4 - 4 \leq 0\\
        &\eta(2-N_{\max})^2 + 4(1-N_{\max}) = \eta(1 - \eta/4)^2 - \eta \leq 0
    \end{align*}
    hence it follows that $\max_{u \in [0, N_{\max}]} f(N_{\max}, u) \leq N_{\max}$ which completes the proof.
    \end{proof}
The previous proposition can be used to show that the individual coordinates themselves are bounded, that they obey a ``sign-stability'' property, and that $\Psi_t$ is decreasing. We show this in the following corollary which restates Corollary \ref{cor:bounded_coordinates} in the main paper.
\begin{corollary}\label{cor:bounded_coordinates_app}
    Under the conditions of Proposition \ref{prop:bounded_iterates_app}, for all times $t$ we have $|c_t(i)| < 1$ and $\sign(c_t(i)) = \sign(c_0(i))$ for all $i \in [n]$. Furthermore, $\Psi_t$ is monotonically decreasing. 
\end{corollary}
\begin{proof}
    By Proposition \ref{prop:bounded_iterates_app} we have $\norm{\bw_t}^2 \leq 1 + \eta / 4$ for all $t$. Define $u_t := \norm{\Pi_{\cB_t}(\bw_t)}^2$ as before.  For any $i \in \cB_t$, the coordinate update is 
    \begin{equation*}
        c_{t+1}(i) = c_t(i)(1 + \eta(2 - u_t - \norm{\bw_t}^2)).
    \end{equation*}
    We have that $\sign(c_{t+1}(i)) = \sign(c_t(i))$ since
    \begin{equation*}
        1 + \eta(2 - u_t - \norm{\bw_t}^2) \geq 1 + \eta(2 - 2\norm{\bw_t}^2) \geq 1 -  \eta^2 / 2 > 0,
    \end{equation*}
    therefore by induction $\sign(c_t(i)) = \sign(c_0(i))$ for all $t$. Furthermore, 
    \begin{equation*}
        |c_{t+1}(i)| = |c_t(i)| (1 + \eta(2 - u_t - \norm{\bw_t}^2)) \leq |c_t(i)| (1 + 2\eta(1 - c_t(i)^2)).
    \end{equation*}
    Using a direct calculation (see Lemma \ref{lem:upper_bound_coord} for details), this implies that if $|c_t(i)| < 1$ then $|c_{t+1}(i)| < 1$, hence by induction $|c_t(i)| < 1$ for all $t$.
    Now consider $j \in [n] \setminus \cB_t$. The coordinate update is
    \begin{equation*}
        c_{t+1}(j) = c_t(j)(1 - \eta u_t)^2.
    \end{equation*}
    Note that these coordinates are mutliplied by a quantity in $(0, 1)$ since
    \begin{align*}
         1 - \eta u_t \leq 1 \text{ and } 1 - \eta u_t
         \geq 1 - \eta \norm{\bw_t}^2
         \geq 1 - \eta(1 + \eta / 4) > 0.  
    \end{align*}
    which easily implies the remaining claims.
\end{proof}
The next two lemmas we show that the iterates can never have small norm and suggest that typically the norm show grow to one. Note that in the proofs of these lemmas we use a slightly modified definition of $u_t$. 
\begin{lemma}\label{lem:small_norm_increases}
    If $\norm{\bw_t}^2 \geq 1 - \eps$ for some $\eps \in (0, 1)$, then $\norm{\bw_{t+1}}^2 \geq \norm{\bw_t}^2 + 4 \eta \eps \sum\limits_{i \in \cB_t} c_t(i)^2$.
    \end{lemma}
    \begin{proof}
    Let $u_t := \sum\limits_{i \in \cB_t} c_t(i)^2$ and $v_t := \norm{\bw_t}^2 - u_t = \sum\limits_{j \in [n] \setminus \cB_t} c_t(j)^2$. Then we can lower bound the increments as follows
    \begin{align*}
        \sum\limits_{i \in \cB_t} c_{t+1}(i)^2 - \sum\limits_{i \in \cB_t} c_t(i)^2
        &= \sum\limits_{i \in \cB_t} c_t(i)^2(1 + \eta(2 - u_t - \norm{\bw_t}^2))^2 - c_t(i)^2\\
        &= u_t(1 + \eta(2 - 2u_t - v_t))^2 - u_t\\
        &= \eta u_t (2 - 2u_t - v_t)(2 + \eta(2- 2u_t -v_t))\\ 
        &= 2\eta u_t(2 - 2u_t - v_t) + \eta^2 u_t(2 - 2u_t - v_t)^2\\
        &\geq 2\eta u_t(2 - 2u_t - v_t),\\
        \sum_{j \in [n] \setminus \cB_t} c_{t+1}(j)^2 - \sum_{j \in [n] \setminus \cB_t} c_t(j)^2
        &= \sum_{j \in [n] \setminus \cB_t} c_{t}(j)^2(1 - \eta u_t)^2 -  c_t(j)^2\\
        &= v_t((1 - \eta u_t)^2 - 1)\\
        &= -\eta v_t u_t(2 - \eta u_t) \geq -2\eta v_t u_t.
    \end{align*}
    Combining the bounds we have
    \begin{align*}
        \norm{\bw_{t+1}}^2 - \norm{\bw_t}^2 &= \qty(\sum\limits_{i \in \cB_t} c_{t+1}(i)^2 - \sum\limits_{i \in \cB_t} c_t(i)^2) + \qty(\sum_{j \in [n] \setminus \cB_t} c_{t+1}(j)^2 - \sum_{j \in [n] \setminus \cB_t} c_t(j)^2)\\
        &\geq 2 \eta u_t (2 - 2u_t - 2v_t)\\
        &\geq 4 \eta u_t (1 - \norm{\bw_t}^2).
    \end{align*}
    The conclusion now follows since by assumption $1 - \norm{\bw_t}^2 \geq \eps$.
    \end{proof}
    
    \begin{lemma}\label{lem:norm_above_one}
    If $\norm{\bw_t}\geq 1$, then $\norm{\bw_{t+1}} \geq 1$.
    \end{lemma}
    \begin{proof}
    Let $N_t := \norm{\bw_t}^2$ and as in the previous lemma define $u_t := \norm{\Pi_{\cB_{t}}(\bw_t)}^2$ and $v_t := N_t - u_t$. Now observe that
    \begin{align*}
        N_{t+1} = \sum\limits_{i \in \cB_t} c_{t+1}(i)^2 + \sum\limits_{j \in [n] \setminus \cB_t} c_{t+1}(j)^2   &= u_t(1 + \eta(2 - 2u_t - v_t))^2 + v_t(1 - \eta u_t)^2\\
        &\geq u_t(1 + 2\eta(2 - 2u_t - v_t)) + v_t(1 - 2 \eta u_t)\\
        &= u_t + v_t + 2 \eta u_t(2 - 2u_t - 2v_t).
    \end{align*}
    The above inequality can be written as
    \[
    N_{t+1} \geq N_t + 4 \eta u_t (1 - N_t).
    \]
    The claim now follows since $N_t + 4 \eta u_t (1 - N_t) \geq 1$. Indeed
    note that
    \[
        N_t + 4 \eta u_t (1 - N_t) \geq 1    
    \]
    if and only if
    \[
    (1 - 4 \eta u_t)(N_t - 1) \geq 0
    \]
    which is true since $N_t \geq 1$ and by Proposition \ref{prop:bounded_iterates_app}
    \[
     4 \eta u_t \leq 4 \eta N_t \leq 4\eta(1 + \eta/4) \leq (4/5) \cdot (1 + 1/20) \leq 1
    \]
    since $\eta \leq 1/5$ by Assumption \ref{ass:init}.
    \end{proof}
    Lastly, we show that $\Phi_t = \norm{\Pi_m(\bw_t)}^2$ is always lower bounded by a constant. 
    \begin{lemma}\label{lem:phi_lower_bound}
        We have $\Phi_t \geq \delta$ for all $t$, where we define the constant 
        \begin{equation}\label{eq:phi_lower_bound}
            \delta :=  \min(\Phi_0, 1 - \Psi_0) > 0.
        \end{equation}
    \end{lemma}
    \begin{proof}
    While $\norm{\bw_t}^2 = \Phi_t + \Psi_t < 1$ we must have $\Phi_t$ increasing since $\norm{\bw_t}^2$ is increasing by Lemma \ref{lem:small_norm_increases} and $\Psi_t$ is decreasing by Corollary \ref{cor:bounded_coordinates_app}, hence $\Phi_t \geq \Phi_0$. If at some point $\norm{\bw_t}^2 \geq 1$, then by Lemma \ref{lem:norm_above_one} for all $t$ thereafter $\Phi_t \geq 1 - \Psi_t \geq 1 - \Psi_0$. Combining these two cases gives that $\Phi_t \geq \delta$ as desired.
    \end{proof}
\subsection{Technical Lemmas}
\begin{lemma}\label{lem:upper_bound_coord}
    Let $f(x) = x(1 + \lambda(1 - x^2))$. If $\lambda \in (0, 1/2]$, then $f(x) \in (0, 1)$ for all $x \in (0, 1)$.
    \end{lemma}
    \begin{proof}
    Computing the derivative $f'(x) = 1 + \lambda - 3\lambda x^2$. Note that $f'(x) > 0$ iff
    \[
    x^2 < \frac{1 + \lambda}{3 \lambda}.
    \]
    For $0 < \lambda \leq 1/2$ we have
    \[
    \frac{1 + \lambda}{3 \lambda} \geq 1,
    \]
    therefore $f'(x) > 0$ for $x \in (0, 1)$. Thus for $x \in (0, 1)$, $0 = f(0) < f(x) < f(1) = 1$.
    \end{proof}
\newpage
\section{General Properties of Minibatch SGD}\label{app:minibatch_sgd}
Now we will present some general properties of minibatch SGD on linear autoencoders. By minibatch SGD, we are referring to the minibatch GD algorithm where at each iteration $t$, the minibatch $\cB_t$ chosen uniformly from the subsets of $[m]$ of size $b$. In contrast, the results in the previous section hold for any sequence of minibatch indices. 

Here and in the next section the maximal index 
\begin{equation}\label{eq:maximal_index}
    i_t^\star := \argmax_{i \in [m]} |c_t(i)|
\end{equation}
will play an important role in the analysis of SGD.  The first result for this section shows that $\psi_t\to 0$. 
\begin{proposition}\label{prop:psi_to_zero}
    As $t\to \infty$, almost surely $\Psi_t \to 0$.
\end{proposition}
\begin{proof}
    Define $u_t := \norm{\Pi_{\cB_t}(\bw_t)}^2$. By Corollary \ref{cor:bounded_coordinates_app}, $\Psi_t$ is decreasing and hence converges to some limiting value $\Psi^\star$. For the sake of contradiction, assume that $\Psi^\star > 0$. Then almost surely $i_t^\star \in \cB_t$ infinitely often. Therefore by Lemma \ref{lem:phi_lower_bound}, $u_t \geq c_t(i_t^\star)^2 \geq \Phi_t / m \geq \delta / m$, infinitely often with $\delta$ defined in Eq.\ (\ref{eq:phi_lower_bound}). Let $T(\varepsilon)$ be a time such that
    \begin{equation*}
        \Psi_T \leq (1 + \varepsilon)\Psi^\star.
    \end{equation*}
    Then almost surely, there exists some $t \geq T$ such that $u_t \geq \Phi_t / m \geq \delta / m$, hence
    \begin{equation*}
        \Psi_{t+1} = \Psi_t(1 - \eta u_t)^2 \leq \Psi_T(1 - \eta \delta / m)^2 \leq (1 + \varepsilon) (1 - \eta \delta / m) \Psi^\star.
    \end{equation*}
If we take $\varepsilon = \eta \delta / m$, then $\Psi_{t+1} < \Psi^\star$ which is a contradiction hence $\Psi^\star = 0$.
\end{proof}
\begin{proposition}\label{prop:liminf_norm}
    Almost surely $\liminf_{t \to \infty} \norm{\bw_t} \geq 1$.
\end{proposition}
\begin{proof}
If $\norm{\bw_t} \geq 1$ at some time $t$, then the claim immediately follows from Lemma \ref{lem:norm_above_one}. Therefore let us assume that $\norm{\bw_t} < 1$ for all $t$. For the sake of contradiction assume that $\liminf\limits_{t \to \infty} \norm{\bw_t} < 1$. Therefore there exists some $\tilde{\varepsilon} > 0$ such that $\norm{\bw_t} \geq 1 - \tilde{\eps}$ at most finitely many times. Since we assumed $\norm{\bw_t} < 1$ for all $t$, in fact there exists $\eps > 0$ such that $\norm{\bw_t}^2 \leq 1 - \varepsilon$ for all $t$. Almost surely there exists a countably infinite set of times $(t_k)_{k=0}^\infty$ such that $i_{t_k} = i_{t_k}^\star$. By Lemma \ref{lem:small_norm_increases} we have that $\norm{\bw_t}$ is monotonically increasing for all $t$ and that for any index $s = t_k$, 
\[
\norm{\bw_{s+1}}^2 - \norm{\bw_s}^2 \geq 4 \eta \varepsilon \sum_{i \in \cB_s} c_t(i)^2 \geq  4 \eta \varepsilon c_t(i_s^\star)^2 \geq 4 \eta \varepsilon \Phi_t / m \geq  4 \eta \varepsilon \delta / m,\] 
where the last inequality is by Lemma \ref{lem:phi_lower_bound} with $\delta$ defined in Eq.\ (\ref{eq:phi_lower_bound}). Hence it is clear that $\norm{\bw_t} \to \infty$ which is a contradiction and so $\liminf\limits_{t \to \infty} \norm{\bw_t} \geq 1$.
\end{proof}
\newpage
\section{Minibatch SGD Convergence}\label{app:sgd_convergence}
In this section we will prove the convergence theorem for minibatch SGD stated in Theorem \ref{thm:sgd_convergence}. Define $\Z_+ := \{0, 1, \ldots\}$ to be the set of nonnegative integers and $\R_+ := [0, \infty)$ to be the set of nonnegative reals. Recall the random variable $R_t$ which is defined as 
\begin{equation}\label{eq:log_ratio_def}
    R_t := \log\qty(\frac{|c_t(i_t^\star)|}{\sum\limits_{\ell \in [m]\setminus \{i_t^\star\}} |c_t(\ell)|}),
\end{equation}
where $i_t^\star$ is defined in Eq.\ (\ref{eq:maximal_index}). A key step will be to prove Proposition \ref{prop:log_ratio_transience} which establishes the transience of the stochastic process $R = (R_t)_{t \in \Z^+}$. From there, we will be able to give the convergence behavior of the iterates $\bw_t$ by invoking results from previous sections. 

\subsection{Transience of $R$}\label{app:log_ratio_transience}
For the purpose of analysing the stochastic process $R$ we can slightly simplify matters and for the proof of Proposition \ref{prop:log_ratio_transience} given in this section assume the following without loss of generality, 
\begin{assumption}\label{ass:pos_init}
    $c_t(\ell) > 0$ for all $\ell \in [n]$ and for all $t$.
\end{assumption}
To see why, observe that for any initialization $(c_0(1), \ldots, c_0(n))$, we can also consider a coupled ``parallel" trajectory induced by running minibatch SGD on the initialization $(\wt{c}_0(1), \ldots, \wt{c}_0(n))$ where $\wt{c}_0(\ell) = c_0(\ell) \cdot \sign(c_0(\ell))$ using the same minibatch sequence. By Corollary \ref{cor:bounded_coordinates_app} it is not hard to see that $|c_t(\ell)| = c_t(\ell) \cdot \sign(c_0(\ell)) = \wt{c}_t(\ell)$ for all $t$, hence $R_t = \wt{R}_t$ where $\wt{R}_t$ is the stochastic process in Eq.\ (\ref{eq:log_ratio_def}) induced by the parallel trajectory. Note that $\wt{c}_t(\ell)$ satisfies Assumptions \ref{ass:init} and \ref{ass:pos_init}, and that $R_t$ is transient if and only if $\wt{R}_t$ is transient, therefore it suffices to analyse trajectories obeying Assumption \ref{ass:pos_init}. 

To establish the transience of the stochastic process $R$ we use a general result from non-homogeneous random walk theory stated in \checked{Proposition~\ref{prop:transience}} which gives general conditions for transience. Recall that $\cB_t \subseteq [m]$ is the selected minibatch indices at time $t$ where where each minibatch is selected at uniform from $\{\cB \subseteq [m]: |\cB| = b\}$. Let $\cF_t = \sigma(\cB_0, \ldots, \cB_t)$ be the sigma-algebra generated by the random minibatch draws $\cB_0, \ldots, \cB_t$. Clearly the stochastic process $R$ is adapted to the filtration $(\cF_t)_{t \in \Z^+}$. 

Note that $c_t(i_t^\star) \geq c_t(\ell)$ for any $\ell \in [m]$ by definition of $i_t^\star$ and so we have the deterministic lower bound $R_t \geq -\log(m)$. Therefore $R$ is a stochastic process on the (translated) half-line. To invoke the theorem directly we will have to verify that $R$ obeys assumptions \ref{enum:irreducibility} and \ref{enum:moment_condition} as well as lower bound the conditional mean increment, that is, show that there exists a function $\lmu_1 : \R \to \R$ such that for all $t \in \Z^+$ 
\[
    \lmu_1(R_t) \leq \E(R_{t+1} - R_t \mid \cF_t),~~\text{a.s.},
\]
 and $\liminf\limits_{r \to \infty} \lmu_1(r) > 0$. For ease of presentation we lower bound the conditional increment first and then verify the assumptions. To this end we will show the following. 
\begin{proposition}\label{prop:increment_lower_bound}
    For all $t \in \Z_+$, there exists $\delta_1 : \R \to \R_+$ and $\delta_2: \R^n \to \R_+$ such that
    \begin{equation*}
        \E(R_{t+1} - R_t \mid \cF_t) \geq \frac{\eta b(m-b)^2}{2m(m-1)^2} - \delta_1(R_t) - \delta_2(\bw_t) - \delta_1(R_t)\delta_2(\bw_t),~~\text{a.s.},
    \end{equation*}
    $\delta_1(r) \to 0$ as $r \to \infty$, and $\delta_2(\bw_t) \to 0$ almost surely as $t \to \infty$.
\end{proposition}
Note that this is actually sufficient for our purposes since we can apply the theorem to the tail process $(R_t)_{t \geq \tau(\eps)}$ where $\tau(\eps)$ defined as the stopping time
\begin{equation*}
    \tau(\eps) := \inf\{\tau \in \Z^+: \delta_2(\bw_t) \leq \eps, \text{ for all } t \geq \tau\},~~\eps = \frac{\eta b(m-b)^2}{4m(m-1)^2},
\end{equation*}
since the proposition gives that for all $t \geq \tau(\eps)$
\begin{equation*}
    \lmu_1(R_t) := \frac{\eta b(m-b)^2}{2m(m-1)^2} - \delta_1(R_t) - \eps - \eps \delta_1(R_t) \leq  \E(R_{t+1} - R_t \mid \cF_t),~~\text{a.s.}, 
\end{equation*}
and that 
\begin{equation*}
    \liminf\limits_{t \to \infty} \lmu_1(r) = \liminf\limits_{t \to \infty} \frac{\eta b(m-b)^2}{2m(m-1)^2} - \delta_1(r) - \eps - \eps \delta_1(r) = \frac{\eta b(m-b)^2}{2m(m-1)^2} - \eps = \eps > 0.
\end{equation*}   
    To obtain the lower bound in Proposition \ref{prop:increment_lower_bound} we will bound the increment at time $t$
    \begin{equation*}
         \Delta_t := R_{t+1} - R_t = (R_{t+1} - R_t) \ind(i_t^\star \in \cB_t) + (R_{t+1} - R_t) \ind(i_t^\star \not\in \cB_t),
    \end{equation*}
    by considering two cases.\:
    \begin{enumerate}
        \item In the first case, $i_t^\star \in \cB_t$ and we will give a positive lower bound on $\Delta_t$.
        \item In the second case, $i_t^\star \not\in \cB_t$ and we will upper bound how negative $\Delta_t$ can be.
    \end{enumerate}
      By averaging over these two cases we will obtain an asymptotically positive lower bound $\lmu_1$ as desired. Now let us introduce some notation that will be useful in this section. Define
      \begin{align*}
        \cJ_t &:= [m] \setminus \{i_t^\star\},&& S_t := \sum\limits_{\ell \in \cJ_t} c_t(\ell)\\
        u_t(\cB) &:= \sum\limits_{\ell \in \cB} c_t(\ell)^2,&&\Psi_t := \sum\limits_{\ell \in [n]\setminus [m]} c_t(\ell)^2
      \end{align*}
      Note that $S_t = c_t(i_t^\star) \cdot \exp(-R_t)$ by definition of $R_t$. We also define the update factors
      \begin{equation*}
        A_t(\cB) = 1 + \eta(2 - u_t(\cB) - \norm{\bw_t}^2),~~ B_t(\cB) = 1 - \eta u_t(\cB).
      \end{equation*}
    Observe that $A_t(\cB)$ and $B_t(\cB)$ are precisely the multiplicative factors such that 
    \begin{align*}
        c_{t+1}(i) &= c_t(i) \cdot A_t(\cB_t) && \text{if}~ i \in \cB_t,\\
         c_{t+1}(j) &= c_t(j) \cdot B_t(\cB_t) && \text{if}~ j \not\in \cB_t.
    \end{align*}
    Denote $\ol{\cB} = [m] \setminus \cB$. Then we define the final set of quantities
    \begin{align*}
        X_t(\cB) &= \sum\limits_{\ell \in \cB} c_t(\ell), &&Y_t(\cB) = \sum\limits_{\ell \in \ol{\cB}} c_t(\ell),\\
        \wt{X}_t(\cB) &= \sum\limits_{\ell \in \cB \setminus \{i_t^\star\}} c_t(\ell), &&\wt{Y}_t(\cB) = \sum\limits_{\ell \in \ol{\cB} \setminus \{i_t^\star\}} c_t(\ell).
    \end{align*}
    Clearly, if $i_t^\star \in \cB$ then $S_t = \wt{X}_t(\cB) + Y_t(\cB)$ and $S_t = X_t(\cB) + \wt{Y}_t(\cB)$ otherwise. For convenience we use $o(R_t)$ and $o(t)$ to make the following substitutions
    \begin{align*}
        o(R_t) &\equiv f(R_t),~~f : \R \to \R_+,~~ \lim\limits_{r \to \infty} f(r) = 0\\
        o(t) &\equiv g(\bw_t),~~g : \R^n \to \R_+,~~ \lim\limits_{t \to \infty}g(\bw_t) = 0,~ \text{a.s.}
    \end{align*}
    For example, we will write $o(R_t)$ in place of $\exp(-R_t)$. We will also use
    \begin{equation*}
        \mathfrak{o}(t, R_t) \equiv o(R_t) + o(t) + o(t)o(R_t)
    \end{equation*}
    to make further substitutions when terms of this form which arise. Before giving the proof of Proposition \ref{prop:increment_lower_bound} we give a lemma which gives asymptotic bounds on the ratio of $A_t(\cB) / B_t(\cB)$ which will important later for lower bounding the conditional mean increment.
    \begin{lemma}\label{lem:ab_ratio_asymptotic}
        Let $\cB \subseteq [m]$ be a set of minibatch indices. If $i_t^\star \in \cB$ then
        \begin{equation*}
            \frac{A_t(\cB)}{B_t(\cB)} \geq \frac{1}{1-\eta} - o(R_t) - o(t) = \frac{1}{1-\eta} - \mathfrak{o}(t, R_t), 
        \end{equation*}
        and if $i_t^\star \not\in \cB$ then 
        \begin{equation*}
            \frac{A_t(\cB)}{B_t(\cB)} \leq 1+ \eta + o(t) + o(R_t) + o(t)o(R_t) = 1 + \eta + \mathfrak{o}(t, R_t).
        \end{equation*}
    \end{lemma}
    \begin{proof}
        To begin with it will be helpful to recall that $c_t(\ell) < 1$ for all $\ell \in [n]$ by Corollary \ref{cor:bounded_coordinates_app} and $\Psi_t = o(t)$ by Proposition \ref{prop:psi_to_zero}, therefore
        \begin{equation*}
            S_t = \sum\limits_{\ell \in \cJ_t} c_t(\ell) = c_t(i_t^\star) \exp(-R_t) \leq \exp(-R_t) = o(R_t)
        \end{equation*}
        and furthermore
        \begin{align*}
            \norm{\bw_t}^2 &= c_t(i_t^\star)^2 + \sum\limits_{\ell \in \cJ_t} c_t(\ell)^2 + \Psi_t\\
            &\leq c_t(i_t^\star)^2 + \sum\limits_{\ell \in \cJ_t} c_t(\ell) + \Psi_t\\
            &\leq 1 + o(R_t) + o(t).
        \end{align*}
        Now let us start with the first inequality we we wish to show. Assume that $i_t^\star \in \cB$. Then,
        \begin{align*}
            \frac{A_t(\cB)}{B_t(\cB)} &= \frac{1 + \eta(2 - u_t(\cB) - \norm{\bw_t}^2)}{1 - \eta u_t(\cB)}\\ 
            &= \frac{1 + \eta(2 - u_t(\cB))}{1 - \eta u_t(\cB)} - \frac{\eta}{1 - \eta u_t(\cB)} \norm{\bw_t}^2\\
            &=\frac{1 + \eta(2 - u_t(\cB))}{1 - \eta u_t(\cB)} - \frac{\eta}{1 - \eta u_t(\cB)} (u_t(\cB) + \norm{\bw_t}^2 - u_t(\cB))\\
            &= \frac{1 + 2 \eta(1 - u_t(\cB))}{1 - \eta u_t(\cB)} - \frac{\eta}{1 - \eta u_t(\cB)}(\norm{\bw_t}^2 - u_t(\cB))\\
            &\geq \frac{1 + 2 \eta(1 - u_t(\cB))}{1 - \eta u_t(\cB)} - \frac{\eta}{1 - 2 \eta}(\norm{\bw_t}^2 - u_t(\cB))
        \end{align*}
        where the last line uses the fact that $u_t(\cB) \leq \norm{\bw_t}^2 \leq 2$ by Proposition \ref{prop:bounded_iterates_app}. Since $i_t^\star \in \cB$,
        \begin{equation*}
            \norm{\bw_t}^2 - u_t(\cB) \leq S_t + \Psi_t = o(R_t) + o(t).
        \end{equation*}
        Combing with the above we have 
        \begin{equation*}
            \frac{A_t(\cB)}{B_t(\cB)} \geq \frac{1 + 2 \eta(1 - u_t(\cB))}{1 - \eta u_t(\cB)} - o(R_t) - o(t).
        \end{equation*}
        Consider the function 
    \[
    f(x) = \frac{1 + 2\eta(1-x)}{1 - \eta x}.
    \]
    A simple computation yields that if $\eta < 1/2$ then $f(x)$ is decreasing for all $x \in \R$ since
    \[
    f'(x) = \frac{\eta(2 \eta - 1)}{(1 - \eta x)^2} < 0.
    \]
    Therefore since $u_t(\cB) \leq \norm{\bw_t}^2 \leq 1 + o(R_t) + o(t)$
    \begin{align*}
        \frac{1 + 2 \eta(1 - u_t(\cB))}{1 - \eta u_t(\cB)} &\geq f(1 + o(R_t) + o(t))\\ 
        &\geq f(1) - o(R_t) - o(t)\\
        &= \frac{1}{1 - \eta} - o(R_t) - o(t).
    \end{align*}
    Therefore combining everything together yields the first desired inequality
    \[
        \frac{A_t(\cB)}{B_t(\cB)} \geq \frac{1}{1 - \eta} - o(R_t) - o(t).
    \]
    For the other inequality, assume now that $i_t^\star \not\in \cB$. Then,
    \begin{align*}
        \frac{A_t(\cB)}{B_t(\cB)} &= \frac{1 + \eta(2 - u_t(\cB) - \norm{\bw_t}^2)}{1 - \eta u_t(\cB)}\\ 
        &= 1 + \eta \frac{2 - \norm{\bw_t}^2}{1 - \eta u_t(\cB)}\\
        &\leq 1 + \eta \frac{2 - \norm{\bw_t}^2}{1 - o(R_t)}
    \end{align*}
    where the last line follows from the fact that $u_t(\cB) \leq S_t = o(R_t)$. Now observe that by Proposition \ref{prop:liminf_norm} we have $\norm{\bw_t}^2 \geq 1 - o(t)$, therefore
    \begin{align*}
        \frac{A_t(\cB)}{B_t(\cB)} &\leq 1 + \eta \frac{1 + o(t)}{1 - o(R_t)}\\
        &\leq 1 + \eta(1 + o(t))(1 + o(R_t))\\
        &= 1 + \eta + o(t) + o(R_t) + o(t)o(R_t).
    \end{align*}
    \end{proof}
    We are now ready to give the proof of Proposition \ref{prop:increment_lower_bound}.
    \begin{proof}[Proposition \ref{prop:increment_lower_bound}]
    First observe that
    \[
    \sum_{\ell \in \cJ_{t+1}} c_{t+1}(\ell) = \sum_{\ell \in [m]} c_{t+1}(\ell) - c_{t+1}(i_{t+1}^\star) \leq \sum_{\ell \in [m]} c_{t+1}(\ell) - c_{t+1}(i_{t}^\star) = \sum_{\ell \in \cJ_{t}} c_{t+1}(\ell)
    \]
    since $c_{t+1}(i_{t+1}^\star) \geq c_{t+1}(i_t^\star)$, hence we have that
    \begin{align}
        R_{t+1} - R_t &= \log\l(\frac{c_{t+1}(i_{t+1}^\star)}{\sum\limits_{\ell \in \cJ_{t+1}} c_{t+1}(\ell)}\r) - \log\l(\frac{c_t(i_t^\star)}{\sum\limits_{\ell \in \cJ_{t}} c_t(\ell)}\r)\nonumber \\ 
        &\geq \log\l(\frac{c_{t+1}(i_{t}^\star)}{\sum\limits_{\ell \in \cJ_{t}} c_{t+1}(\ell)}\r) - \log\l(\frac{c_t(i_t^\star)}{\sum\limits_{\ell \in \cJ_{t}} c_t(\ell)}\r).\label{eq:delta_lower_bound_app}
    \end{align}
    We start with the case $i_t^\star \in \cB_t$. Starting from Eq.\ (\ref{eq:delta_lower_bound_app}), we have that on the event $i_t^\star \in \cB_t$,
    \begin{align}
        R_{t+1}-R_t &\geq \log\l(\frac{c_{t+1}(i_{t}^\star)}{\sum\limits_{\ell \in \cJ_t} c_{t+1}(\ell)}\r) - \log\l(\frac{c_t(i_t^\star)}{\sum\limits_{\ell \in \cJ_t} c_t(\ell)}\r) \nonumber\\
        &= \log\l(\frac{c_{t}(i_{t}^\star) A_t(\cB_t)}{A_t(\cB_t) \wt{X}_t(\cB_t) + B_t(\cB_t) Y_t(\cB_t)}\r) - \log\l(\frac{c_t(i_t^\star)}{S_t}\r)\nonumber\\ 
        &= \log\l(\frac{A_t(\cB_t) S_t}{A_t(\cB_t)(S_t - Y_t(\cB_t)) + B_t(\cB_t)Y_t(\cB_t)}\r)\nonumber\\
        &= \log\l(\frac{S_t}{S_t - (1 - B_t(\cB_t) / A_t(\cB_t)) \cdot Y_t(\cB_t)}\r).\label{eq:delta_intermediate_lower_bound}    
    \end{align}
    By Lemma \ref{lem:ab_ratio_asymptotic},
    \begin{equation*}
        \frac{B_t(\cB_t)}{A_t(\cB_t)} \leq 1 - \eta + \mathfrak{o}(t, R_t),
    \end{equation*}
    hence the above yields
    \begin{align*}
        (R_{t+1} - R_t) \cdot \ind(i_t^\star \in \cB_t) \geq \log\l(\frac{S_t}{S_t - [\eta - \mathfrak{o}(t, R_t)] \cdot Y_t(\cB)}\r) \cdot \ind(i_t^\star \in \cB_t).
    \end{align*}
   Now we can apply Jensen's inequality and obtain that
   \begin{equation*}
    \E(R_{t+1} - R_t \mid i_t^\star \in \cB_t, \cF_{t-1}) \geq \log\qty(\frac{S_t}{S_t - [\eta - \mathfrak{o}(t, R_t)] \cdot \E(Y_t(\cB) \mid i_t^\star \in \cB_t, \cF_{t-1})}),
   \end{equation*}
   Defining $\ol{\cB}_t := [m] \setminus \cB_t$, we can compute 
   \begin{align*}
    \E(Y_t(\cB) \mid i_t^\star \in \cB_t, \cF_{t-1}) &= \E\qty(\sum\limits_{\ell \in [m]} \mathbbm{1}(\ell \in \ol{\cB}_t) \cdot c_t(\ell) \mathrel{\Big|} i_t^\star \in \cB_t, \cF_{t-1})\\
    &= \sum\limits_{\ell \in [m] } \Pr(\ell \in \ol{\cB}_t \mid i_t^\star \in \cB_t) \cdot c_t(\ell)
   \end{align*}
   Recall that $\cB_t$ is uniformly chosen from the set of subsets of $[m]$ of size $b$. Therefore conditional on $i_t^\star \in \cB_t$, $\ol{\cB}_t$ is uniformly chosen from the set of subsets of $[m] \setminus \{i_t^\star\}$ of size $m - b$, hence $\Pr(\ell \in \ol{\cB}_t \mid i_t^\star \in \cB_t)$ is $(m-b)/(m-1)$ if $\ell \neq i_t^\star$ and $0$ otherwise. Therefore
   \begin{align*}
       \E(Y_t(\cB) \mid i_t^\star \in \cB_t, \cF_{t-1}) &= \sum\limits_{\ell \in [m] } \Pr(\ell \in \ol{\cB}_t \mid i_t^\star \in \cB_t) \cdot c_t(\ell)\\
    &= \frac{m - b}{m - 1} \sum\limits_{\ell \in [m] \setminus \{i_t^\star\}} c_t(\ell) = \frac{m - b}{m - 1} S_t,
   \end{align*}
   which then leads to
   \begin{align*}
    \E(R_{t+1} - R_t \mid i_t^\star \in \cB_t, \cF_{t-1}) &\geq \log\qty(\frac{S_t}{S_t - [\eta - \mathfrak{o}(t, R_t)] \cdot  \frac{m - b}{m-1} \cdot S_t})\\
    &= \log\qty(\frac{1}{1 - \eta \cdot \frac{m-b}{m-1}})- \mathfrak{o}(t, R_t).
   \end{align*}
Now consider the case where $i_t^\star \not\in \cB$. Again using Eq.\ (\ref{eq:delta_lower_bound_app}) we have that on this event,
\begin{align*}
    -(R_{t+1} - R_t) &\leq  -\log\l(\frac{c_{t+1}(i_{t}^\star)}{\sum\limits_{\ell \in \cJ_t} c_{t+1}(\ell)} \r) + \log\l(\frac{c_t(i_t^\star)}{\sum\limits_{\ell \in \cJ_t} c_t(\ell)}\r)\\
    &= \log\l(\frac{\wt{Y}_t(\cB_t) B_t(\cB_t) + X_t(\cB_t)A_t(\cB_t)}{c_{t}(i_{t}^\star) B_t(\cB_t)} \r) + \log\l(\frac{c_t(i_t^\star)}{S_t}\r)\\
    &= \log\l(\wt{Y}_t(\cB_t) + X_t(\cB_t) \frac{A_t(\cB_t)}{B_t(\cB_t)} \r) - \log(S_t).
\end{align*}
By Lemma \ref{lem:ab_ratio_asymptotic} we have that if $i_t^\star \not\in \cB_t$ then
\begin{equation*}
    \frac{A_t(\cB_t)}{B_t(\cB_t)} \leq 1 + \eta + \mathfrak{o}(t, R_t).
\end{equation*}
Hence we can bound the above as
\begin{align*}
    -(R_{t+1} - R_t) &= \log\l(\wt{Y}_t(\cB_t) + X_t(\cB_t) \cdot [1 + \eta + \mathfrak{o}(t, R_t)] \r) - \log(S_t)\\
    &= \log\l(\wt{Y}_t + X_t(\cB_t) + X_t(\cB_t) \cdot [\eta + \mathfrak{o}(t, R_t)] \r) - \log(S_t)\\
    &= \log\l(S_t + X_t(\cB_t) \cdot [\eta + \mathfrak{o}(t, R_t)]\r) - \log(S_t)\\
    &= \log\l(1 + \frac{X_t(\cB_t)}{S_t} \cdot [\eta + \mathfrak{o}(t, R_t)]\r)\\ 
    &\leq  \frac{X_t(\cB_t)}{S_t} \cdot [\eta + \mathfrak{o}(t, R_t)]. 
\end{align*}
Thus we have shown that 
\begin{equation*}
    -(R_{t+1} - R_t) \cdot \ind(i_t^\star \not\in \cB_t) \leq \frac{X_t(\cB_t)}{S_t} \cdot [\eta + \mathfrak{o}(t, R_t)] \cdot \ind(i_t^\star \not\in \cB_t).
\end{equation*}
Therefore by negating and taking expectations on both sides
\begin{equation*}
    \E(R_{t+1} - R_t \mid i_t^\star \not\in \cB, \cF_{t-1}) \geq -(\eta + \mathfrak{o}(t, R_t)) \cdot \frac{\E(X_t(\cB) \mid i_t^\star \not\in \cB_t, \cF_{t-1})}{S_t}.
\end{equation*}
We can compute the conditional expectation
\begin{align*}
    \E(X_t(\cB) \mid i_t^\star \not\in \cB_t, \cF_{t-1}) &= \E\qty(\sum\limits_{\ell \in [m] } \mathbbm{1}(\ell \in \cB_t) \cdot c_t(\ell) \mathrel{\Big|} i_t^\star \not\in \cB_t, \cF_{t-1} )\\
    &= \sum\limits_{\ell \in [m] } \Pr(\ell \in \cB_t \mid i_t^\star \not\in \cB_t) \cdot c_t(\ell)\\
    &= \frac{b}{m-1} \sum\limits_{\ell \in [m] \setminus \{i_t^\star\} } c_t(\ell) = \frac{b}{m-1}S_t,
\end{align*}
where we used the fact that conditional on $i_t^\star \not\in \cB_t$, $\cB_t$ is uniformly chosen from the set of subsets of $[m] \setminus \{i_t\}$ of size $b$.
This then gives 
\begin{align*}
    \E(R_{t+1} - R_t \mid i_t^\star \not\in \cB, \cF_{t-1}) &\geq -(\eta + \mathfrak{o}(t, R_t)) \frac{b}{m-1}  \frac{S_t}{S_t} = - \eta \frac{b}{m-1} - \mathfrak{o}(t, R_t).
\end{align*}
Therefore combining the two cases using the law of total probability gives
\begin{align*}
    \E(R_{t+1} - R_t \mid \cF_{t-1}) &= \Pr(i_t^\star \in \cB_t) \cdot \E(R_{t+1} - R_t \mid i_t^\star \in \cB_t, \cF_{t-1})\\ 
    &\quad+ \Pr(i_t^\star \not\in \cB_t) \cdot \E(R_{t+1} - R_t \mid i_t^\star \not\in \cB_t, \cF_{t-1})\\
    &\geq \frac{b}{m}\log\qty(\frac{1}{1 - \eta \cdot \frac{m - b}{m-1}}) - \eta \frac{m-b}{m} \frac{b}{m-1} - \mathfrak{o}(t, R_t)\\
    &\geq \eta \frac{b}{m} \frac{m - b}{m-1}  \qty(1 + \frac{\eta}{2}\frac{m-b}{m-1})  -  \eta\frac{m-b}{m} \frac{b}{m-1}  - \mathfrak{o}(t, R_t)\\
    &= \frac{\eta b(m-b)^2}{2m(m-1)^2} - \mathfrak{o}(t, R_t),
\end{align*}
where the second inequality used that
\begin{equation*}
\log\qty(\frac{1}{1-x}) \geq x \cdot (1 + x/2),~~\text{for all } x \in (0, 1).
\end{equation*}
\end{proof}

Now it remains to verify Assumptions \ref{enum:irreducibility} and \ref{enum:moment_condition}. Let us first consider \ref{enum:irreducibility}. Essentially \ref{enum:irreducibility} will hold because, as suggested in the proof of the previous Proposition \ref{prop:increment_lower_bound}, $R_t$ increases if $i_t^\star \in \cB_t$. We will show that in this case $R_t$ will increase by at least a constant amount for any time $t$. Hence $R_t$ can grow to an arbitrarily large value with constant probability. Here constant probability means that the probability is independent of the past, but potentially dependent on the desired target value, which is what is required by the condition in \ref{enum:irreducibility}. We now show this holds formally.
\paragraph{Proof \ref{enum:irreducibility} holds} 
We will uses parts of the proof of Proposition \ref{prop:increment_lower_bound} and use the same notation. If $i_t^\star \in \cB_t$ then from Eq.\ (\ref{eq:delta_intermediate_lower_bound})
\begin{equation*}
    R_{t+1} - R_t \geq \log\l(\frac{S_t}{S_t - (1 - B_t(\cB_t) / A_t(\cB_t)) \cdot Y_t(\cB_t)}\r).
\end{equation*}
Using the fact that $Y_t(\cB_t) \leq S_t$ we then have
\begin{equation*}
    \Delta_t(\cB) \geq \log\l(\frac{S_t}{S_t - (1 - B_t(\cB_t) / A_t(\cB_t)) \cdot S_t}\r) = \log\l(\frac{\cA_t(\cB_t)}{B_t(\cB_t)}\r).
\end{equation*}
In the proof of the proposition we used an asymptotic lower bound on $A_t(\cB_t) / B_t(\cB_t)$ from Lemma \ref{lem:ab_ratio_asymptotic}, however here we will use the simpler non-asymptotic bound
\begin{align*}
    \frac{A_t(\cB_t)}{B_t(\cB_t)} &= \frac{1 + \eta(2 - u_t(\cB_t) - \norm{\bw_t}^2)}{1 - \eta u_t(\cB_t)}\\
    &\geq 1 + \eta \frac{2- \norm{\bw_t}^2}{1 - \eta u_t(\cB_t)}\\
    &\geq 1 + \eta(1 - \eta / 4) \geq 1 + \frac{19}{20}\eta,
\end{align*} 
which just follows since $0 \leq u_t(\cB_t) \leq \norm{\bw_t}^2 \leq 1 + \eta/4$ by Proposition \ref{prop:bounded_iterates_app} and $\eta \leq 1/5$ by Assumption \ref{ass:init}. Thus we have the lower bound
\begin{equation*}
    R_{t+1} - R_t \geq \log\l(1 + \frac{19}{20}\eta\r).
\end{equation*}
Hence we have shown that if $i_t^\star \in \cB_t$ then $R_{t+1} - R_t \geq \delta$ for $\delta := \log(1 + (19/20)\eta)$. Now we can easily show that \ref{enum:irreducibility} holds. For any $T \in \Z^+$ consider a sequence of minibatches $(\cB_s)_{s=0}^T$. For any $y \in (0, \infty)$ we can take $v(T) = \ceil{\delta^{-1}(y + \log(m))}$ so that if $i_T^\star \in \cB_t$ for $T \leq t \leq v(T)$ then $R_{T + v(T)} \geq y$ since
\begin{align*}
    R_{T + v(T)} \geq R_T + \delta v(T) \geq -\log(m) + \delta \cdot [\delta^{-1}(y + \log(m))] = y.
\end{align*}
Furthermore this occurs with probability $(b/m)^{\ceil{\delta^{-1}(y + \log(m))}} > 0$.

\paragraph{Proof \ref{enum:moment_condition} holds} We now show that the process $(R_t)_{t \in \Z^+}$ has bounded increments. For convenience define $\cJ_t := [m] \setminus \{i_t^\star\}$. By definition
\begin{align*}
     R_{t+1} - R_t  &= \log\l(\frac{c_{t+1}(i_{t+1}^\star)}{\sum\limits_{\ell \in \cJ_{t+1}} c_{t+1}(\ell)}\r) - \log\l(\frac{c_t(i_t^\star)}{\sum\limits_{\ell \in \cJ_{t}} c_t(\ell)}\r)\\
     &= \log\l(\frac{c_{t+1}(i_{t+1}^\star)}{c_t(i_t^\star)} \frac{\sum\limits_{\ell \in \cJ_{t}} c_t(\ell)}{\sum\limits_{\ell \in \cJ_{t+1}} c_{t+1}(\ell)}\r).
\end{align*}
It then suffices to show that the quantity $I$ defined as
\begin{equation*}
    I := I_1\cdot I_2,~~I_1 = \frac{c_{t+1}(i_{t+1}^\star)}{c_t(i_t^\star)}, I_2 =\frac{\sum\limits_{\ell \in \cJ_{t}} c_t(\ell)}{\sum\limits_{\ell \in \cJ_{t+1}} c_{t+1}(\ell)},
\end{equation*}
lies in a time-independent compact subinterval of $(0, +\infty)$. Let us define the following constants $\beta$, $\gamma$
\begin{equation}\label{eq:beta_gamma_const}
    \beta = 1 - 2 \eta \in (0, 1),~~ \gamma = \frac{1 - 2\eta}{1 + 2 \eta} \in (0, 1).
\end{equation}
We will show that
\begin{align}
    \beta &\leq I_1 \leq 1/\beta\label{eq:first_term_bound}\\
    \gamma \beta &\leq I_2 \leq 1/\beta \label{eq:second_term_bound}
\end{align}
Note that for any $\ell \in [m]$, if we define $u_t = \norm{\Pi_{\cB}(\bw_t)}^2$ then
\[
    \min\{1 - \eta u_t, 1 + \eta(2 - u_t - \norm{\bw_t}^2)\} \leq \frac{c_{t+1}(\ell)}{c_t(\ell)} \leq \max\{1 - \eta u_t, 1 + \eta(2 - u_t - \norm{\bw_t}^2)\}.
\]
By Proposition \ref{prop:bounded_iterates_app} since $0 \leq u_t \leq \norm{\bw_t}^2 \leq 2$,
\begin{align*}
    \min\{1 - \eta u_t, 1 + \eta(2 - u_t - \norm{\bw_t}^2)\} &\geq 1 - 2 \eta \\
    \max\{1 - \eta u_t, 1 + \eta(2 - u_t - \norm{\bw_t}^2)\} &\leq \max\{1, 1 + 2 \eta\} = 1 + 2 \eta.
\end{align*}
Therefore for $\beta$ defined as in Eq.\ (\ref{eq:beta_gamma_const}) 
\[
\frac{c_{t+1}(\ell)}{c_t(\ell)} \in [\beta, 1/\beta]
\]
for any $\ell \in [m]$. From this it easily follows $I_1 \in [\beta, 1 / \beta]$ as claimed in Eq.\ (\ref{eq:first_term_bound}) since
\[
\frac{c_{t+1}(i_{t+1}^\star)}{c_t(i_t^\star)} \leq \frac{c_{t+1}(i_{t+1}^\star)}{c_t(i_{t+1}^\star)} \leq 1/\beta,~~ \frac{c_{t+1}(i_{t+1}^\star)}{c_t(i_t^\star)} \geq \frac{c_{t+1}(i_{t}^\star)}{c_t(i_t^\star)} \geq \beta.
\]
Now let us consider the term $I_2$. If $i_{t+1}^\star = i_t^\star$ then it is easy to see that
\begin{equation}\label{eq:I2_case_one}
    I_2 = \frac{\sum\limits_{\ell \in \cJ_{t}} c_t(\ell)}{\sum\limits_{\ell \in \cJ_{t+1}} c_{t+1}(\ell)} = \frac{\sum\limits_{\ell \in \cJ_{t}} c_t(\ell)}{\sum\limits_{\ell \in \cJ_{t}} c_{t+1}(\ell)} \in [\beta, 1/\beta].
\end{equation}
    
Now consider the case when $i_{t+1}^\star \neq i_t^\star$. We claim that is can only happen if $i_{t+1}^\star \in \cB_t$ and $i_t \not\in \cB_t$. To see why, let $A = c_{t+1}(i_{t+1}^\star) / c_t(i_t^\star)$ and $B = c_{t+1}(i_t^\star) / c_t(i_t^\star)$. If this were not true then 
\begin{equation*}
    \frac{c_{t+1}(i_{t+1}^\star)}{c_{t+1}(i_t^\star)} = \frac{c_{t}(i_{t+1}^\star)}{c_t(i_t^\star)} \frac{A}{B} \leq \frac{c_{t}(i_{t+1}^\star)}{c_t(i_t^\star)} \leq 1
\end{equation*}
because $A, B \in \{1 + \eta(2 - u_t - \norm{\bw_t}^2), 1 - \eta u_t\}$ and $A / B > 1$ only if $A = 1 + \eta(2 - u_t - \norm{\bw_t}^2)$ and $B = 1 - \eta u_t$, which is exactly when $i_{t+1}^\star \in \cB_t$ and $i_t \not\in \cB_t$. In this case 
\begin{equation*}
    1 \leq  \frac{c_{t+1}(i_{t+1}^\star)}{c_{t+1}(i_t^\star)} = \frac{c_{t}(i_{t+1}^\star)}{c_t(i_t^\star)} \frac{1 + \eta(2 - u_t - \norm{\bw_t}^2)}{1 - \eta u_t} \leq \frac{c_{t}(i_{t+1}^\star)}{c_t(i_t^\star)} \frac{1 + 2 \eta}{1 - 2 \eta}
\end{equation*}
from which it follows that
\begin{equation*}
    c_{t}(i_{t+1}^\star) \geq c_t(i_t^\star) \frac{1 - 2 \eta}{1 + 2 \eta} = \gamma c_t(i_t^\star).
\end{equation*}
Now observe that $I_2$ lies in the interval
\begin{equation}\label{eq:I2_case_two}
    I_2 = \frac{\sum\limits_{\ell \in \cJ_{t}} c_t(\ell)}{\sum\limits_{\ell \in \cJ_{t+1}} c_{t+1}(\ell)} \in [\beta I_2', (1/\beta) \cdot I_2'],
\end{equation}
where we define the term 
\[
I_2' :=
\frac{\sum\limits_{\ell \in \cJ_{t}} c_t(\ell)}{\sum\limits_{\ell \in \cJ_{t+1}} c_{t}(\ell)} = \frac{\sum\limits_{\ell \in [m] \setminus \{i_t^\star, i_{t+1}^\star\}} c_t(\ell) + c_t(i_{t+1}^\star)}{\sum\limits_{\ell \in [m] \setminus \{i_t^\star, i_{t+1}^\star\}} c_t(\ell) + c_t(i_t^\star)}.
\]
It is clear that $I_2' \leq 1$ because $c_t(i_t^\star) \geq c_t(i_{t+1}^\star)$ and since $c_t(i_{t+1}^\star) \geq \gamma c_t(i_t^\star)$ with $\gamma \in (0, 1)$,
\begin{align*}
I_2' &= \frac{\sum\limits_{\ell \in [m] \setminus \{i_t^\star, i_{t+1}^\star\}} c_t(\ell) + c_t(i_{t+1}^\star)}{\sum\limits_{\ell \in [m] \setminus \{i_t^\star, i_{t+1}^\star\}} c_t(\ell) + c_t(i_t^\star)}\\
&\geq \frac{\sum\limits_{\ell \in [m] \setminus \{i_t^\star, i_{t+1}^\star\}} c_t(\ell) + \gamma c_t(i_{t}^\star)}{\sum\limits_{\ell \in [m] \setminus \{i_t^\star, i_{t+1}^\star\}} c_t(\ell) + c_t(i_t^\star)} \geq \gamma.
\end{align*}
Thus we see that $I_2' \in [\gamma, 1]$. From Eq.\ (\ref{eq:I2_case_two}) it follows that in this case $I_2 \in [\gamma \beta, 1 / \beta]$. Combining with Eq.\ (\ref{eq:I2_case_one}) yields the desired bound on $I_2$ in Eq.\ (\ref{eq:second_term_bound}). Since the bounds on $I_1$ and $I_2$ imply that $I \in [\gamma \beta^2, 1 / \beta^2]$ since this implies
\[
|R_{t+1} - R_t| \leq \max(|\log(\gamma \beta^2)|, |\log(1 / \beta^2)|).
\]
\subsection{Proof of Theorem \ref{thm:sgd_convergence}}\label{app:sgd_thm_proof}
Now we are ready to give the convergence results for minibatch SGD in Theorem \ref{thm:sgd_convergence}.
\begin{proof}
    From Proposition \ref{prop:log_ratio_transience} we know that $R_t \to \infty$ almost surely, that is the ratio of $|c_t(i_t^\star)|$ to $S_t := \sum_{\ell \in \cJ_t} |c_t(\ell)|$ goes to infinity, where $\cJ_t := [m] \setminus \{i_t^\star\}$. Since $|c_t(i_t^\star)| \leq 1$, we know that $S_t \to 0$. We now show that $|c_t(i_t^\star)| \to 1$. With $\Psi_t$ as defined as in Eq.\ (\ref{eq:psi_def}) and using $|c_t(\ell)| < 1$ for all $\ell \in [n]$, we have
    \begin{align*}
        1 \geq c_t(i_t^\star)^2 &= \norm{\bw_t}^2 - \sum\limits_{\ell \in \cJ_t} c_t(\ell)^2 - \Psi_t\\
        &\geq \norm{\bw_t}^2 - |c_t(i_t^\star)| \exp(-R_t) - \Psi_t\\
        &\geq \norm{\bw_t}^2 - \exp(-R_t) - \Psi_t.
    \end{align*}
Now since $\liminf\limits_{t \to \infty} \norm{\bw_t} \geq 1$ by Proposition \ref{prop:liminf_norm} and $\Psi_t \to 0$ by Proposition \ref{prop:psi_to_zero}, 
since $R_t$ is transient 
we can see by taking $t \to \infty$ that indeed $|c_t(i_t^\star)| \to 1$. 

Now it remains to show that eventually $i_t^\star$ becomes constant.  Intuitively, this is true because the fact that $c_t(i_t^\star) \to 1$ while $\max_{\ell \neq i} c_t(\ell)^2 \to 0$ means that the only way for $i_t^\star$ to be non-constant would be for gradient descent to rapidly move all of the mass from one coordinate to another, but this is not possible since the gradient norm goes to zero as we show in Lemma \ref{lem:gradient_norm_bound}.   More formally, 
for any $\eps > 0$, take $T := T(\eps)$ large enough such that $\max_{\ell \neq i_t^\star} |c_t(\ell)| \leq \eps$ and $|c_t(i_t^\star) - 1| \leq \eps$ for all $t \geq T$. By Lemma \ref{lem:gradient_norm_bound},
\[
    \sup_{t \geq T} \max_{\ell \in [n]} |c_{t+1}(\ell) - c_t(\ell)| = O(\eps).
\]
Let $i^\star := i_T^\star$. We will prove that for $\eps$ small enough, $i_t^\star = i^\star$ for all $t \geq T$. For the sake of contradiction, let $t > T$ be the first time such that $i_t^\star \neq i^\star$. From the above we must have
\begin{align*}
    \max_{\ell \neq i^\star} c_t(\ell) &\leq \max_{\ell \neq i^\star} c_{t-1}(\ell) + O(\eps) \leq O(\eps)\\
    c_t(i^\star) &\geq c_{t-1}(i^\star) - O(\eps) \geq 1 - O(\eps)
\end{align*}
Therefore for $\eps$ small enough $i^\star = \argmax_{\ell \in [m]} \abs*{c_t(\ell)}$ which is a contradiction. Thus we have $i_t^\star = i^\star$ for $t \geq T$ and so $|c_t(i^\star)| \to 1$. Furthermore, by Corollary \ref{cor:bounded_coordinates_app} we have that $\sign(c_t(i^\star)) = \sign(c_0(i^\star))$, hence $\bw_t \to \sign(c_0(i^\star)) \cdot \ba_i$ which is what we wished to show.
\end{proof}
\begin{lemma}\label{lem:gradient_norm_bound}
Consider a trajectory $(\bw_t)_{t \in \Z^+}$ of minibatch GD. Assume that for some $t$, there exists $i \in [m]$ and $\eps > 0$, such that 
\[|c_t(i) - 1| \leq \eps \text{ and } |c_t(j)| \leq \eps~ \text{ for all } j \in [n]\setminus \{i\}.
\]
Then as $\eps \to 0^+$,
\[
    \max_{\ell \in [n]}~|c_{t+1}(\ell) - c_t(\ell)| \leq O(\eps).
\]
\end{lemma}
\begin{proof}
    Define $u_t = \norm{\Pi_{\cB_t}(\bw_t)}^2$. Consider $\ell \not\in \cB_t$,
    then
    \[
        |c_{t+1}(\ell) - c_t(\ell)| = \eta |c_t(\ell)| u_t \leq \eta m \eps = O(\eps).
    \]
    Now consider $\ell \in \cB_t$. Then for $\eps$ small enough,
    \begin{align*}
         2 - u_t - \norm{\bw_t}^2 &\geq 2 - 2\norm{\bw_t}^2\\
         &\geq  2 - 2(1+\eps)^2 - 2(n-1)\eps^2\\
         &= -4\eps - 2n\eps^2,\\
        2 - u_t - \norm{\bw_t}^2 &\leq 2 - (1 - \eps)^2 - (1-\eps)^2\\
         &= 4\eps - 2\eps^2.
    \end{align*}
    Hence $|2 - u_t - \norm{\bw_t}^2| = O(\eps)$ and
    \[
        |c_{t+1}(\ell) - c_t(\ell)| = \eta|c_t(\ell)||(2 - u_t -\norm{\bw_t}^2)| = O(\eps).
    \]
\end{proof}

\section{Full-batch Gradient Descent Convergence}\label{app:gd_convergence}
In this section we will give the proof of Theorem \ref{thm:gd_convergence} which gives the convergence of full-batch gradient descent. As a reminder, we will make Assumption \ref{ass:init} throughout, that is we assume that $\norm{\bw_0} < 1$ and $\eta \leq 1/5$, where $\eta := \alpha / m$. Let us recall some definitions. Define the vector of correlations 
\begin{equation*}
    \bc_t = (\ip{\bw_t}{\ba_1}, \ldots, \ip{\bw_t}{\ba_n}) \in \R^n
\end{equation*}
Furthermore we will define
\begin{equation*}
    \Phi_t := \snorm{\Pi_m(\bw_t)}^2 = \sum\limits_{i \in [m]} c_t(i)^2,~~~~ \Psi_t := \norm{\bw_t}^2 - \snorm{\Pi_{m}(\bw_t)}^2 = \sum\limits_{j \in [n] \setminus [m]} c_t(j)^2.
\end{equation*}
From Lemma \ref{lem:coordinate_updates} we can write the full-batch gradient update as follows
\begin{align*}
    c_{t+1}(i) &= c_t(i) +  \eta c_t(i) (2 - 2 \Phi_t - \Psi_t) && i \in [m],\\
    c_{t+1}(j) &= c_t(j) - \eta c_t(j) \Phi_t && j \in [n] \setminus [m].
\end{align*}
To reduce notational clutter in the following we will sometimes suppress the time index $t$ and write for example $c_i := c_t(i)$, $c_i' := c_{t+1}(i)$, and $\Delta c_i = c_i' - c_i$. 
For example, we can write the full-batch update at time $t$ as follows
\begin{align}
    \Delta c_i &= \eta c_i(2 - 2 \Phi - \Psi) && i \in [m] \label{eq:gd_pos},\\ 
    \Delta c_j &= -\eta c_j \Phi && j \in [n] \setminus [m] \label{eq:gd_neg}.
\end{align}
We will first show that the dynamics obey an important invariant due to the symmetry of the updates which is not true when the batch size $b < m$.
\begin{proposition}\label{prop:gd_invariant}
    For all $i \in [m]$ and for all $t \in \Z^+$,
    \begin{equation*}
        \frac{c_t(i)}{c_0(i)} = \sqrt{\frac{\Phi_t}{\Phi_0}}.
    \end{equation*}
\end{proposition}
\begin{proof}
    Define the quantity,
\begin{equation*}
    \Gamma_t = \eta(2 - 2 \Phi_t - \Psi_t).
\end{equation*}
From Eq.\ (\ref{eq:gd_pos})
\[
\Delta c_i = \eta c_i \Gamma,~~i\in [m].
\]
or equivalently
\[
c_{t+1}(i) = c_t(i)(1 + \Gamma_t),~~i\in [m].
\]
Unrolling the updates over $t$ yields
\[
    c_t(i) = c_0(i) \prod\limits_{k=0}^{t-1}(1 + \Gamma_k), ~~~~i \in [m].
\]
By squaring and summing both sides of the above over $i \in [m]$ we see
\[
   \Phi_t = \Phi_0 \l[\prod\limits_{k=0}^{t-1}(1 + \Gamma_k)\r]^2
\]
It immediately follows that
\[
\frac{c_t(i)}{c_0(i)} = \sqrt{\frac{\Phi_t}{\Phi_0}},~~~~i \in [m].
\]
\end{proof}
Therefore to obtain the convergence behavior of $c_t(i)$ for $i \in [m]$ it suffices to understand the limit of $\Phi_t$. The same proof given in Proposition \ref{prop:psi_to_zero} will give $\Psi_t \to 0$ for full-batch GD, which implies $c_t(j) \to 0$ for all $j \not\in [n] \setminus [m]$, but this fact will emerge in the proofs from this section anyways. Thus, to analyse the convergence of $\bw_t$ it suffices to analyse the limits of $\Phi_t$ and $\Psi_t$. We can easily write the update equations for the dynamics of $\Phi_t$ and $\Psi_t$ solely in terms of these two quantities.
\begin{lemma}\label{lem:norm_updates}
    The updates for $\Phi_t$ and $\Psi_t$ are given by
    \begin{align*}
        \Delta \Phi &= 2\eta \Phi (2 - 2 \Phi - \Psi) + \eta^2 \Phi (2 - 2 \Phi - \Psi)^2 \\
        \Delta \Psi &= -2\eta \Phi \Psi + \eta^2 \Phi^2 \Psi.
    \end{align*}
    \end{lemma}
    \begin{proof}
    This follow from straight-forward calculations
    \begin{align*}
        \Delta \Phi = \Phi' - \Phi &= \sum_{i \in S} (c_i')^2 - c_i^2\\
        &= \sum_{i \in S} (c_i' - c_i)(c_i' + c_i)\\
        &= \sum_{i \in S} \eta c_i (2 - 2 \Phi - \Psi)(2c_i + \eta c_i(2 - 2\Phi - \Psi)) \\
        &= 2\eta \sum_{i \in S} c_i^2 (2 - 2 \Phi - \Psi) + \eta^2 \sum_{i \in S} c_i^2 (2 - 2 \Phi - \Psi)^2 \\
        &= 2\eta \Phi (2 - 2 \Phi - \Psi) + \eta^2 \Phi (2 - 2 \Phi - \Psi)^2,
    \end{align*}
    and similarly
    \begin{align*}
        \Delta \Psi = \Psi' - \Psi &= \sum_{j \in S^c} (c_j')^2 - c_j^2\\
        &= \sum_{j \in S^c} (c_j' - c_j)(c_j' + c_j)\\
        &= \sum_{j \in S^c} -\eta c_j \Phi(2c_j - \eta c_j \Phi) \\
        &= -2\eta \sum_{j \in S^c} c_j^2 \Phi + \eta^2 \sum_{j \in S^c} c_j^2 \Phi^2 \\
        &= -2\eta \Phi \Psi + \eta^2 \Phi^2 \Psi.
    \end{align*}
    \end{proof}
The next proposition establishes the asymptotic convergence of $\Phi_t$ and $\Psi_t$. In particular we will show the following
\begin{proposition}\label{prop:phi_psi_convergence}
    $\Phi_t$ monotonically increases to $1$ and $\Psi_t$ monotonically decreases to $0$.
\end{proposition}
To show the above proposition, we will first show that the following quantity
\[
    \cN_t = \Phi_t + \frac{5}{8}\Psi_t
\]
remains bounded above by one, from which the proposition will easily follow. 
\begin{lemma}\label{lem:weight_norm_less_one}
If $\cN_t < 1$, then $\cN_{t+1} < 1$.
\end{lemma}
\begin{proof}
Consider the update at time $t$. By definition
\[
    \Delta \cN = \Delta \Phi + \frac{5}{8} \Delta \Psi.
\]
From Lemma \ref{lem:norm_updates} we have that
\begin{align*}
    \Delta \Phi &= 2\eta \Phi (2 - 2 \Phi - \Psi) + \eta^2 \Phi (2 - 2 \Phi - \Psi)^2 \\
    \Delta \Psi &= -2\eta \Phi \Psi + \eta^2 \Phi^2 \Psi.
\end{align*}
We will show that
\begin{align*}
    \Delta \Phi &\leq 5 \eta \Phi (1 - \Phi - \Psi/2),\\
    \Delta \Psi &\leq -\eta \Phi \Psi.
\end{align*}
Since $\Phi + (5/8) \Psi < 1$ and $\Phi, \Psi \geq 0$ it follows that 
\[
    0 \leq \Phi < 1 \text { and } 0 < 1 - \Phi - \Psi / 2 \leq 1.
\]
Furthermore since $\eta \leq 1/5$, we can bound $\Delta \Phi$ as follows
\begin{align*}
    \Delta \Phi &= 2\eta \Phi (2 - 2 \Phi - \Psi) + \eta^2 \Phi (2 - 2 \Phi - \Psi)^2\\
    &= 4\eta \Phi (1 - \Phi - \Psi / 2) + 4\eta (1 - \Phi - \Psi / 2) \cdot [\eta \Phi (1 - \Phi - \Psi / 2)]\\
    &\leq 4\eta \Phi (1 - \Phi - \Psi / 2) + \frac{4}{5}\eta (1 - \Phi - \Psi / 2)\\ 
    &\leq 5\eta \Phi (1 - \Phi - \Psi / 2).
\end{align*}
Similarly, for $\Delta \Psi$ we have
\begin{align*}
    \Delta \Psi &= -2\eta \Phi \Psi + \eta^2 \Phi^2 \Psi\\
    &= -2\eta \Phi \Psi + \eta \Phi \Psi [\eta \Phi]\\
    &\leq -2\eta \Phi \Psi + \frac{1}{5} \eta \Phi \Psi \leq -\eta \Phi \Psi.
\end{align*}
Now observe that from the previous inequalities
\begin{align*}
    \Delta \cN &= \Delta \Phi + \frac{5}{8} \Delta \Psi\\ 
    &\leq 5\eta \Phi (1 - \Phi - \Psi/2) - \frac{5}{8} \eta \Phi \Psi\\
    &= 5\eta \Phi (1 - (\Phi + 5\Psi/8))\\
    &\leq 5 \eta (\Phi + 5\Psi/8) (1 - (\Phi + 5\Psi/8))\\
    &= 5\eta \cN(1 - \cN).
\end{align*}
Since $\eta \leq 1/5$ by Assumption \ref{ass:init}, it follows $\cN_{t+1} < 1$ by a simple calculation (see Lemma \ref{lem:bounded_by_one}).
\end{proof}
We are now ready to give the proof of Proposition \ref{prop:phi_psi_convergence}.
\begin{proof}[Proposition \ref{prop:phi_psi_convergence}]
We will first show that $\Phi_t \to 1$. Since $\cN_0 < 1$ by Assumption \ref{ass:init}, from Lemma \ref{lem:weight_norm_less_one} it follows by induction that $\cN_t = \Phi_t + (5/8)\Psi_t < 1$ for all $t$. Let us now consider the updates at a particular time. We have that
\[
2 - 2 \Phi - \Psi \geq 2(1-\Phi) - \frac{8}{5} (1 - \Phi) = \frac{2}{5}(1 - \Phi) > 0.
\]
 Recall from Lemma \ref{lem:norm_updates} that 
\begin{align*}
    \Delta \Phi &= 2 \eta \Phi(2 - 2 \Phi - \Psi) + \eta^2 \Phi(2 - 2 \Phi - \Psi)^2\\
    &\geq 2 \eta \Phi(2 - 2\Phi - \Psi).
\end{align*}
Thus we see that since $\Delta \Phi \geq 0$, $\Phi_t$ is monotonically increasing and
\begin{align*}
    \Delta \Phi &\geq 2 \eta \Phi(2 - 2\Phi - \Psi) \\
    &\geq (2\eta) \cdot \frac{2}{5}(1 - \Phi)\\
    &= \frac{4}{5} \eta \Phi(1 - \Phi)\\
    &\geq \frac{4}{5} \eta  \Phi_0 \cdot (1 - \Phi).
\end{align*}
Thus, by a simple calculation (see Lemma \ref{lem:convergence_to_one}), for all $t$ we have
\[
0 \leq 1 - \Phi_t \leq (1 - \Phi_0) \cdot \exp(-\kappa t)
\]
where $\kappa := (4/5)\eta \Phi_0 > 0$, hence $\Phi_t \to 1$ as desired. From Corollary \ref{cor:bounded_coordinates_app} we know $\Psi_t$ is monotonically decreasing. By the squeeze theorem it is easy to see that $\Psi_t \to 0$, since $\Phi_t + (5/8)\Psi_t < 1$ implies that
\[
0 \leq \Psi_t \leq \frac{8}{5}(1 - \Phi_t).
\]
\end{proof}
Now we are ready to give the proof of Theorem \ref{thm:gd_convergence}.
\begin{proof}[Theorem \ref{thm:gd_convergence}]
Recall from Proposition \ref{prop:gd_invariant} that
\begin{equation*}
    \frac{c_t(i)}{c_0(i)} = \sqrt{\frac{\Phi_t}{\Phi_0}}
\end{equation*}
for all $i \in [m]$ and $t \in \Z^+$. From Proposition \ref{prop:phi_psi_convergence} we have $\Phi_t \to 1$, hence
\[
c_t(i) \to \frac{c_0(i)}{\sqrt{\Phi_0}},~~~~i \in [m],
\]
as well as $\Psi_t \to 0$, from which it is clear that $c_t(j) \to 0$ for $j \in [n] \setminus [m]$. Therefore we see
\[
\bw_t \to \frac{1}{\sqrt{\Phi_0}} \sum_{i \in [m]} c_0(i) \ba_i
\]
as we wished to show.
\end{proof}

We now give the proof of Corollary \ref{cor:low_correlation}.
For convenience, we will say an event occurs with high probability (w.h.p) if it occurs with probability at least $1 - O(m^{-1})$. 
\begin{proof}[Corollary \ref{cor:low_correlation}]
Define $\overline{\bw}_t = \bw_t / \norm{\bw_t}$. By Theorem~\ref{thm:gd_convergence} we have that
    \[
        \lim\limits_{t \to \infty}\overline{\bw}_t =  \frac{\Pi_m(\bw_0)}{\snorm{\Pi_m(\bw_0)}},~~~~ \Pi_m(\bw_0) = \sum_{i \in [m]} \ip{\bw_0}{\ba_i} \ba_i.
    \]
    Therefore our goal is to show that w.h.p
    \[
       \max_{i \in [m]}~ \abs*{\ip{\ba_i}{\lim\limits_{t \to \infty}\overline{\bw}_t}} = \frac{1}{\sqrt{\Phi}} \max_{i \in [m]} |c_0(i)| = O\qty(\frac{\sqrt{n \log m}}{m}),~~\Phi := \sum\limits_{i \in [m]} c_0(i)^2.
    \]
    We will do so by bounding both $\Phi$ and $\max_{i \in [m]} \abs*{c_0(i)}$ w.h.p. 
    Since $c_0(i) \simiid \mathcal{N}(0, \sigma_{\init}^2 / n)$
    it follows that 
    \[
        \Phi = \sum_{i\in [m]} c_0(i)^2 \sim \sigma_{\init}^2 / n\cdot \chi^2(m),
    \]
    where $\chi^2(m)$ denotes a chi-squared random variable with $m$ degrees of freedom. The standard tail bound in Lemma \ref{lem:chi-square} implies that w.h.p, 
    \[
    \Phi \geq \frac{\sigma_{\init}^2}{n} \cdot \frac{m}{4}.
    \]
    Now by a standard inequality for the maximum absolute value of independent Gaussians stated in Lemma \ref{lem:max_gaussians}, we have that w.h.p, 
    \[
        \max_{i\in [m]} \abs*{c_0(i)} \leq 3\sigma_{\init}\sqrt{\frac{\log (2m)}{n}}.
    \]
    Combining everything together then yields  w.h.p    
    \[
    \frac{1}{\sqrt{\Phi}} \cdot \max_{i \in [m]} \abs*{c_0(i)} \leq \frac{2 \sqrt{n}}{\sigma_{\init} \sqrt{m}} \cdot 3\sigma_{\init} \sqrt{\frac{\log(2m)}{n}} = O\qty(\sqrt{\frac{\log m}{m}}).
    \]
    Now the desired claim follows since $m = \Theta(n)$. 
\end{proof}
    

\subsection{Technical Lemmas}
\begin{lemma}\label{lem:convergence_to_one}
Consider a sequence $\{x_t\}_{t \in \N}$ which satisfies
\[
x_{t+1} - x_t \geq c_t(1 - x_t)
\]
for all $t \in \N$, where $c_t \in (0, 1]$ and $x_0 \leq 1$. Then
\[
1 - x_t \leq \prod\limits_{i = 1}^t (1 - c_i) (1 - x_0) \leq \exp\l(-\sum\limits_{i = 1}^t c_i\r) (1 - x_0)
\]
\end{lemma}
\begin{proof}
Rearranging 
\[
x_{t+1} - x_t \geq c_t(1 - x_t)
\]
yields
\[
(1 - x_{t+1}) \leq (1 - c_t) (1 - x_t)
\]
hence unrolling the recursion yields 
\[
1 - x_t \leq \prod\limits_{i = 1}^t (1 - c_i) (1 - x_0)
\]
and then the inequality $1 - x \leq e^{-x}$ yields
\[
\prod\limits_{i = 1}^t (1 - c_t) (1 - x_0) \leq \exp\l(-\sum\limits_{i = 1}^t c_i\r) (1 - x_0).
\]
\end{proof}

\begin{lemma}\label{lem:bounded_by_one}
Let $\{x_t\}_{t \in \N}$ be a sequence such that $x_0 < 1$ and
\[
x_{t+1} - x_t \leq \lambda x_t(1 - x_t)
\]
for $\lambda \leq 1$. Then $x_t < 1$ for all $t \in \N$.
\end{lemma}
\begin{proof}
Assume the statement is true for $t \leq T$. Observe that the function
\[
f(x) = (1 + \lambda)x - \lambda x^2
\]
has derivative
\[
f'(x) = 1 + \lambda - 2 \lambda x
\]
hence $f$ is strictly increasing on the interval $(-\infty, 1]$ and $f(1) = 1$. Therefore since $x_T \in [0, 1)$, we have that $x_{T + 1} \leq f(x_{T}) < 1$ completing the claim.
\end{proof}

\subsection{Concentration Inequalities}
 \begin{lemma}[Chi-square Tail Bound]\label{lem:chi-square}
    If $X \sim \chi^2(k)$ then for all $t \in (0, 1)$,
    \[
        \Pr[X \leq k(1-t)] \leq \exp(-kt^2/8).
    \]
    \end{lemma}
    \begin{lemma}[Chernoff Bound]\label{lem:chernoff}
    Let $X = \sum_{i=1}^n X_i$ where $X_i = 1$ with probability $p_i$ and $X_i = 0$ with probability $1 - p_i$, and all $X_i$ are independent. Let $\mu = \E(X) = \sum_{i=1}^n p_i$. Then
    \[
    \Pr(X \leq (1 - \delta) \mu) \leq \exp(-\mu \delta^2 / 2)
    \]
    for all $\delta \in (0, 1)$.
    \end{lemma}
    \begin{lemma}[Maximum of Gaussians]\label{lem:max_gaussians}
    Let $X_1, \ldots, X_n \simiid \cN(0, \sigma^2)$. Then,
    \[
        \Pr\l(\max_{i \in [n]} |X_i| - \sqrt{2 \sigma^2 \log (2n)} \geq t\r) \leq \exp\l(\frac{-t^2}{2 \sigma^2}\r).
    \]
    \end{lemma}

\section{Cyclic SGD Convergence}\label{app:csgd_convergence}
\subsection{Dynamics of Cyclic SGD}\label{sec:csgd_dynamics}
First let us recall the setting of Theorem \ref{thm:csgd_convergence}. We assume that $m = n = 2$, that is our dataset $\cD = \{\ba_0, \ba_1\}$ where $\ba_i \in \R^2$. At each time step $t$ we process example $\bx_t$ where $\bx_t = \ba_{t \tmod 2}$ and $t \tmod 2$ is $0$ when $t$ is even and $1$ when $t$ is odd.
Let $y_t := \sip{\bw_t}{\ba_0}$ and $z_t := \sip{\bw_t}{\ba_1}$. We assume that $y_0 \geq z_0 > 0$. 
From Lemma \ref{lem:coordinate_updates} it follows that the dynamics are given by
\begin{align*}
    y_{t + 1} &= y_t(1 + \eta(2 - 2y_t^2 - z_t^2))\\
    z_{t + 1} &= z_t(1 - \eta y_t^2)
\end{align*}
for $t \tmod 2 = 0$ and for $t \tmod 2 = 1$
\begin{align*}
    y_{t + 1} &= y_t(1 - \eta z_t^2)\\
    z_{t + 1} &= z_t(1 + \eta(2 - 2z_t^2 - y_t^2)).
\end{align*}
For convenience we let $\twostep : \R^2 \to \R^2$ denote the function which gives the two-step (epoch) update
\[
    (y_{t+2}, z_{t+2}) = \twostep(y_t, z_t),~~t \tmod 2 = 0.
\]
Note that by Proposition \ref{prop:bounded_iterates_app} and \ref{cor:bounded_coordinates_app}, if $y_0^2 + z_0^2 < 1$ and $\eta \leq 1/4$ then for all $t$,
\[
    (y_t, z_t) \in \{(y, z): 0 < y, z < 1,~~ y^2 + z^2 \leq 1 + \eta / 4\}.
\]
We will make use of the following definitions
\begin{itemize}
    \item Define the potential function $V(y, z) = z/y$, which gives the relative alignment (compare with Eq.\ (\ref{eq:log-ratio})). We will show that the potential is always decreasing by at least a constant each epoch in order to prove that $V(y_t, z_t) \to 0$.
    \item Define $\decrease = \{(y, z) \in (0, 1)^2 : V(F(y, z)) - V(y, z) < 0\}$ as the set of points where the potential strictly decreases after an epoch. We will show $V$ decreases each epoch by proving that the iterates are always in this set (in particular the subset $\cA$ defined below) at the start of the epoch.
    \item Define $\attract = \{(y, z) \in (0, 1)^2 : y \geq z > 0, y^2 + z^2 \leq 1 + \eta / 4\}$. We will show that $\attract \subseteq \decrease$ and that $\attract$ is an invariant set under $F$, i.e. $(y, z) \in \attract$ implies $F(y, z) \in \attract$. 
\end{itemize}
\subsection{Proof of Theorem \ref{thm:csgd_convergence}}

We will consider the subsequence of even iterates $(y_{2t}, z_{2t})$ for $t = 0, 1, \ldots$ Let us recall the sets $\decrease$ and $\attract$ and the epoch update function $F$ defined in Appendix \ref{sec:csgd_dynamics}. In Proposition \ref{prop:attracting_set} we show that $\attract \subseteq \decrease$. By the definition of $\decrease$ it is easy to see that $\attract$ is invariant under $\twostep$, that is if $(y, z) \in \attract$ then $\twostep(y, z) \in \attract$. Since by assumption $(y_0, z_0) \in \attract$, this will imply that $(y_{2t}, z_{2t}) \in \attract$ for all $t$ and that $V(y_{2t}, z_{2t})$ is strictly decreasing. Thus by the monotone convergence theorem there exists some $V_\star \in [0, +\infty)$ such that $V(y_{2t}, z_{2t}) \to V_\star$.

We claim that $V_\star = 0$. For the sake of contradiction assume that $V_\star > 0$. Let $N_t = y_t^2 + z_t^2$. Since by assumption $N_0 < 1$, from Lemmas \ref{lem:small_norm_increases}, \ref{lem:norm_above_one} and Proposition \ref{prop:bounded_iterates_app}, we have that for all $t$, 
\[
0 < N_0 \leq N_{2t} \leq 1 + \eta/4.
\]
Since $V_\star \leq z_{2t} / y_{2t} \leq z_{0}/y_0 \leq 1$, the sequence $(y_{2t}, z_{2t})$ lies in the annulus $\cK_1$ where 
\[
\cK_1 = \{(r \cos \theta, r \sin \theta) : \tan(\theta) \in [V_\star, 1],  r \in [N_0, 1 + \eta/4]\}.
\] 
Note that $\cK_1 \subseteq \decrease$ is a compact set. Thus we have a contradiction by Proposition \ref{prop:exit} and therefore $V_\star = 0$. 
Now we show that $\lim (y_{2t}, z_{2t}) = (1, 0)$. By Corollary \ref{cor:bounded_coordinates_app}
\[
y_{2t}^2 = N_{2t} - (z_{2t}/y_{2t})^2 \cdot y_{2t}^2 \geq N_{2t} - (z_{2t}/y_{2t})^2
\]
which implies that
\begin{align*}
\liminf y_{2t}^2 &\geq \liminf N_{2t} - \lim (z_{2t}/y_{2t})^2\\
&= \liminf N_{2t} - V_\star\\
&= \liminf N_{2t} \geq 1,
\end{align*}
where the last inequality follows from Proposition \ref{prop:liminf_norm} because $i_{2t} = i_{2t}^\star = 0$ for all $t$. Since $y_{2t}^2 \leq 1$ we have $\limsup y_{2t}^2 \leq 1$. Therefore $\lim y_{2t} = 1$ and $\lim z_{2t} = \lim y_{2t} \cdot (z_{2t} / y_{2t}) = 0$. We have shown that the even subsequence converges to the desired limit point. Now invoking Lemma \ref{lem:gradient_norm_bound} is is easy to see that $(y_{t}, z_{t}) \to (1, 0)$ as desired.

\subsection{Auxiliary Results}
\begin{proposition}\label{prop:exit}
    Let $\{x_t\}_{t=0}^\infty$ be a sequence in $\R^n$ such that there exists continuous $F: \R^n \to \R^n$ and $x_{t+1} = F(x_t)$ for all $t = 0, 1 \ldots$ Assume there exists a function $V : \R^n \to \R$ that is continuous on a compact subset $\cK \subseteq \R^n$ such that for all $x \in \cK$, $V(F(x)) - V(x) < 0$. Then there exists $t_0 \in \N$ such that $x_{t_0} \not\in \cK$. 
\end{proposition}
\begin{proof}
    For the sake of contradiction assume that $x_t \in \cK$ for all $t$. Define the quantity 
    \[
        \varepsilon := \sup\{V(F(x)) - V(x) : x \in \cK\}.
    \]
    By the continuity of $V$ and $F$ and the compactness of $\cK$, it follows that $\varepsilon < 0$. Therefore for any $T$, 
    \begin{align*}
        \inf_{x \in \cK} V(x) &\leq V(x_T)\\
        &= V(x_0) + \sum\limits_{t = 0}^{T-1} V(x_{t+1}) - V(x_t)\\
        &= V(x_0) + \sum\limits_{t = 0}^{T-1} V(F(x_t)) - V(x_t)\\ 
        &\leq V(x_0) + \varepsilon T.
    \end{align*}
    However, the inequality
    \[
        \inf_{x \in \cK} V(x) \leq V(x_0) + \varepsilon T
    \]
     cannot hold since the left-hand side is finite and the right-hand side approaches negative infinity as $T \to \infty$.
\end{proof}

\begin{proposition}\label{prop:attracting_set}
    The set $\attract = \{(y, z) : y \geq z > 0, y^2 + z^2 \leq 1 + \eta / 4\} \subseteq \decrease$.
\end{proposition}
\begin{proof}
Let $y_0 = r \cos \theta$ and $z_0 = r \sin \theta$ with $\theta \in [0, \pi / 2]$. Consider fixing $r$ and varying $\theta$. Observe that
\begin{align*}
    V(F(r \cos \theta, r \sin \theta)) - V(r \cos \theta, r \sin \theta) &= \tan \theta \cdot \l(\frac{(1 - \eta y_0^2)}{(1 + \eta(2 - 2y_0^2-z_0^2))}\frac{(1 + \eta(2 - 2z_1^2 - y_1^2)}{(1 - \eta z_1^2)} - 1\r).
\end{align*}
Therefore $(y_0, z_0) \in \decrease$ iff
the following inequality holds
\[
\frac{(1 - \eta y_0^2)}{(1 + \eta(2 - 2y_0^2-z_0^2))} \leq \frac{(1 - \eta z_1^2)}{(1 + \eta(2 - 2z_1^2 - y_1^2)}
\]
or equivalently
\[
(1 - \eta y_0^2)(1 + \eta(2 - 2z_1^2 - y_1^2) \leq (1 - \eta z_1^2)(1 + \eta(2 - 2y_0^2-z_0^2)).
\]
Let us observe that we can write the following terms solely as a function of $r$ and $y_0$.
\begin{align*}
    z_0^2 &= r - y_0^2\\
    z_1^2 &= z_0^2(1 - \eta y_0^2)^2 = (r - y_0^2)(1 - \eta y_0^2)^2 \\
    y_1^2 &= y_0^2(1 + \eta(2 - 2y_0^2 - z_0^2)^2 = y_0^2(1 + \eta(2 - r - y_0))^2.
\end{align*}
Letting $y = y_0$ for convenience and substituting into the above inequality, it is equivalent to
\[
f(y;r) - g(y;r) \leq 0
\]
where
\begin{align*}
    f(y; r) &= (1 - \eta y^2)(1 + \eta[2 - 2(r - y^2)(1 - \eta y^2)^2 - y^2(1 + \eta(2 - r - y^2))^2])\\
    g(y; r) &= (1 - \eta (r - y^2)(1 - \eta y)^2)(1 + \eta(2 - r - y^2)).
\end{align*}
By Lemma \ref{lem:rotation_derivative}
\[
\dv{}{y} f(y;r) - g(y;r) \leq 0.
\]
Recalling $y = r \cos \theta$, by the chain rule
\begin{equation}\label{eq:gap_decreasing_with_angle}
    \dv{}{\theta} [f(y(\theta);r) - g(y(\theta);r)] = \dv{}{y} [f(y;r) - g(y;r)] \dv{y}{\theta} = \dv{}{y} [f(y;r) - g(y;r)] (-r \sin \theta) \geq 0.
\end{equation}
As $\cos(\pi / 4) = \sin(\pi/4) = 1 / \sqrt{2}$, Lemma \ref{lem:diag_in_decrease} states that if $r \leq \sqrt{1 + \eta / 4}$ then $(r \cos \pi/4, r \sin \pi / 4) \in \decrease$, that is
\[
f(r \cos \pi / 4; r) - g(r \cos \pi / 4; r) < 0.
\]
From Eq.\ (\ref{eq:gap_decreasing_with_angle}) for $0 \leq \psi \leq \pi / 4$ 
\[
f(r \cos \psi; r) - g(r \cos \psi; r) \leq f(r \cos \pi / 4; r) - g(r \cos \pi / 4; r) < 0
\]
hence $(r \cos \psi, r \sin \psi) \in \decrease$. Since 
\[
\attract = \{(r \cos \psi, r \sin \psi) : r^2 \leq 1 + \eta / 4, \psi \in [0, \pi / 4]\}
\]
this proves the claim.
\end{proof}

\begin{lemma}\label{lem:diag_in_decrease}
If $0 < y^2 \leq \frac{1}{2} (1 + \eta / 4)$ and $\eta \leq 1/4$, then $(y, y) \in \decrease$.
\end{lemma}
\begin{proof}
Observe that 
\[
(y, y) \in \decrease \iff \frac{z_2}{y_2} - 1 > 0 \iff y_2 - z_2 > 0.
\]
We will explicitly show that the last inequality for $y$ such that $y^2 \leq (1 + \eta / 4) / 2$. We have that
\begin{align*}
    y_1 &= y(1 + \eta(2 - 3y^2))\\
    z_1 &= y(1 - \eta y^2)
\end{align*}
Therefore
\[
y_1 = (1 + \delta)z_1,~~~~ \delta = \frac{2 \eta(1 - y^2)}{1 - \eta y^2}.
\]
Thus we have that
\begin{align*}
    y_2 - z_2 &= y_1(1 - \eta z_1^2) - z_1(1 + \eta(2 - 2 z_1^2 - y_1^2))\\
    &= \delta z_1 - \eta(1 + \delta)z_1^3 - 2 \eta z_1 + 2\eta z_1^3 + \eta(1 + \delta)^2 z_1^3\\
    &= z_1(\delta - 2 \eta) + \eta z_1^3(2 + \delta + \delta^2)
\end{align*}
Substituting and factoring yields
\[
    z_1(\delta - 2 \eta) + \eta z_1^3(2 + \delta + \delta^2) = 2\eta z_1 y^2\l(\frac{(\eta - 1)}{1 - \eta y^2} + (1 - \eta y^2)^2\l(1 + \frac{ \eta(1 - y^2)}{1 - \eta y^2} + \frac{2 \eta^2(1 - y^2)^2}{(1 - \eta y^2)^2}\r) \r)
\]
Letting $w = 1 - \eta y^2$ we thus $y_2 - z_2 \geq 0$ iff
\begin{align*}
    \frac{\eta - 1}{w} + w^2\l(1 + \frac{\eta(1 - y^2)}{w} + \frac{2 \eta^2 (1 - y^2)^2}{w^2}\r) > 0
\end{align*}
Letting $b = 1 - \eta$, it follows that $\eta(1 - y^2) = w - b$ and so the above is equivalent to
\begin{align*}
    -\frac{b}{w} + w^2\l(1 + \frac{w-b}{w} + \frac{2 (w-b)^2}{w^2}\r) > 0
\end{align*}
which after clearing denominators and grouping terms is equivalent to
\[
4w^3 - 5bw^2 + 2wb^2 - b > 0.
\]
The claim then follows from Lemma \ref{lem:diag_w_b_positive}.
\end{proof}


\subsection{Technical Lemmas}

\begin{lemma}\label{lem:diag_w_b_positive}
Assume $\eta \leq 1/4$ and $2y^2 \leq 1 + \eta / 4 $. Let $w = 1 - \eta y^2$ and $b = 1 - \eta$. Then
\[
4w^3 - 5bw^2 + 2wb^2 - b \geq 0.
\]
\end{lemma}
\begin{proof}
Let $f(w, b) = 4w^3 - 5bw^2 + 2wb^2 - b$. Since by assumption $y^2 \leq (1 + \eta/4)/2$, 
\[
w \geq 1 - \eta / 2 - \eta^2 / 8 = \frac{1}{8}(-b^2 + 6b + 3).
\]
Let us call
\[
w_{\min} = \frac{1}{8}(-b^2 + 6b + 3).
\]
Observe that for $w \in [b, 1]$
\[
\dv{}{w} f(w, b) = 12w^2 - 10bw + 2b^2 \geq 14b^2 - 10b \geq 0 
\]
since
\[
14b^2 - 10 b \geq 0 \iff b \geq 5/7 \iff \eta \leq 2/7
\]
which is true since by assumption $\eta \leq 1/4 \leq 2 / 7$.
Further, note that $w_{\min} \geq b$ since 
\[ 8(w_{\min}-b) = -b^2 + 6b + 3 - 8b = -(b^2 + 2b - 3) = -(b+3)(b-1),\]
and $8(w_{\min}-b) >0$ for $b\in [0,1]$.  We thus have,
\[
\inf_{w \in [w_{\min}, 1]} f(w, b) = f(w_{\min}, b).
\]
Using Mathematica to simplify
\[
f(w_{\min}, b) = -\frac{1}{128}(b-1)^2(b^4 - 6b^3 - 2b^2 + 2b - 27).
\]
Since $b \in [0, 1)$ 
\[
b^4 - 6b^3 - 2b^2 + 2b - 27 \leq 1 + 2 - 27 < 0
\]
hence $f(w_{\min}, b) > 0$.
\end{proof}

\begin{lemma}\label{lem:rotation_derivative}
Assume $r \leq 1 + \eta / 4$ is a constant. Define the following functions of $y \in [0, 1]$.
\begin{align*}
    f(y; r) &= (1 - \eta y^2)(1 + \eta[2 - 2(r - y^2)(1 - \eta y^2)^2 - y^2(1 + \eta(2 - r - y^2))^2])\\
    g(y; r) &= (1 - \eta (r - y^2)(1 - \eta y)^2)(1 + \eta(2 - r - y^2)).
\end{align*}
Then the following is true
\[
\dv{}{y} f(y;r) - g(y;r) \leq 0
\]
\end{lemma}
\begin{proof}
Making the substitution $w = 1 - \eta y^2 \iff \eta y^2 = 1 - w$ we have
\begin{align*}
    f(w; r) &= (1 - \eta y^2)(1 + \eta[2 - 2(r - y^2)(1 - \eta y^2)^2 - y^2(1 + \eta(2 - r - y^2))^2])\\
    &= (1 - \eta y^2)(1 + 2\eta + 2( \eta y^2 - \eta r)(1 - \eta y^2)^2 - \eta y^2(1 - \eta y^2 + \eta(2-r ))^2])\\
    &= w[1 + 2 \eta + 2w^2(1 - w - \eta r) + (w - 1)(w + \eta(2 - r))^2]. \\
    g(w; r) &= (1 - \eta (r - y^2)(1 - \eta y)^2)(1 + \eta(2 - r - y^2))\\
    &= (1 + (\eta y^2 - \eta r)(1 - \eta y)^2)(1 - \eta y^2 + \eta(2 - r))\\
    &= (1 + w^2(1 - w - \eta r))(w + \eta(2 - r)).
\end{align*}
Using Mathematica we have that
\begin{align*}
    \dv{}{w} f(w;r) - g(w;r) &= \eta[\eta(2 - r)(r-2+4w) + 6(3-2r)w^2 - 6(2-r)w + 2]\\
    &= \eta[p(r, w) + q(r, w)].
\end{align*}
where
\begin{align*}
    p(r, w) &= \eta(2 - r)(r-2+4w)\\
    q(r, w) &= 6(3-2r)w^2 + 6(r-2)w + 2.
\end{align*}
We now show that $p(r, w) \geq 0$ and $q(r, w) \geq 0$. 
\paragraph{Proof that $p(r, w) \geq 0$}\mbox{}\\ Note that $r \leq 1 + \eta / 4 \leq 2$ hence $2 - r \geq 0$ and since $y^2 \leq 1$ it follows that $w \geq 1 - \eta$, hence $4w \geq 4(1 - \eta) \geq 3$
hence $(r - 2 + 4w) \geq r + 1 \geq 0$ since $r \geq 0$. Therefore $p(r, w) = \eta(2 - r)(r-2+4w) \geq 0$.
\paragraph{Proof that $q(r, w) \geq 0$}\mbox{}\\ Note that we can write
\[
q(r, w) = 6(3-2r)w^2 + 6(r-2)w + 2 = 6rw(1-2w) + s(w)
\]
for some function $s$ of w. 
Since $1 - 2w \leq 1 - 2(1 - \eta) = -1 + 2\eta \leq 0$ it follows that $q$ is decreasing in $r$ therefore
$q(r, w) \geq q(1+\eta/4) \geq q(1 + \eta, w)$. We can lower bound this as follows, using $\eta \leq 1/4$
\begin{align*}
    q(1 + \eta, w) &= 6(3 - 2(1 + \eta))w^2 - 6(1-\eta)w + 2\\
    &\geq 3w^2 - \frac{9}{2} w + 2\\
    &\geq 3(3/4)^2 - (9/2)(3/4) + 2 = 5/16 \geq 0.
\end{align*}
Therefore we have shown that 
\[
\dv{}{w} f(w;r) - g(w;r) \geq 0
\]
and since $w = 1 - \eta y^2$ by the chain rule this implies that
\[
\dv{}{y} f(y;r) - g(y;r) \leq 0.
\]
\end{proof}
\newpage
\section{Loss Landscape}\label{app:loss_landscape}
\subsection{Proof of Theorem \ref{thm:global_minima}}

    It is clear that for minimizing the loss objective we can restrict our consideration to $\bw \in \Span(\ba_1, \ldots, \ba_m)$. Let us define $c_i := \ip{\bw}{\ba_i}$ for $i \in [m]$. Then
    \begin{align*}
        \cL(\bw; \dataset) &= \frac{1}{2m} \sum\limits_{i=1}^m \norm{\ba_i - \bw \phi(\ip{\bw}{\ba_i})}^2\\
        &= \frac{1}{2m} \sum\limits_{i=1}^m \norm{\ba_i - \phi(c_i) \sum\limits_{j=1}^m c_j \ba_j}^2 \\
        &= \frac{1}{2m} \l(\sum\limits_{i=1}^m (1 - c_i \phi(c_i))^2 + \sum\limits_{j \neq i} c_j^2 \phi(c_i)^2\r).
    \end{align*}
    Define the quantity $B := \sum\limits_{j=1}^m c_j^2$. 
    Then recalling that the activation $\phi(t) = \max(t, 0)$
    \begin{align*}
        \cL(\bw; \dataset) &= \frac{1}{2m} \l(\sum\limits_{i=1}^m (1 - c_i \phi(c_i))^2 + \phi(c_i)^2(B - c_i^2)\r)\\
        &= \frac{1}{2m} \l(m - 2 \sum\limits_{i=1}^m c_i \phi(c_i) + \sum\limits_{i=1}^m c_i^2 \phi(c_i)^2 + \sum\limits_{i=1}^m \phi(c_i)^2(B - c_i^2)\r)\\
        &= \frac{1}{2m} \l(m - 2 \sum\limits_{i=1}^m c_i \phi(c_i) + B \sum\limits_{i=1}^m \phi(c_i)^2\r)\\
        &= \frac{1}{2m} \l(m - 2 \sum\limits_{i=1}^m c_i \phi(c_i) + \sum\limits_{i=1}^m c_i^2 \sum\limits_{i=1}^m \phi(c_i)^2\r).
    \end{align*}
    Therefore to find a minimizer it suffices to minimize the quantity
    \begin{equation}\label{eq:global_minimizer_quantity}
        - 2 \sum\limits_{i=1}^m c_i \phi(c_i) + \sum\limits_{i=1}^m c_i^2 \sum\limits_{i=1}^m \phi(c_i)^2.
    \end{equation}
    If we define
    \[
        P = \sum_{i : c_i > 0} c_i^2,~~~~ N = \sum_{i : c_i < 0} c_i^2.
    \]
    then Eq.\ (\ref{eq:global_minimizer_quantity}) can be rewritten as
    \[
    -2P + P(P + N) = P^2 - 2P + PN
    \]
    where $P, N \geq 0$. It is easy to see that the minimum of this quantity is achieved precisely when $P = 1$, $N = 0$, which is what we wished to prove.

\subsection{One-sided Derivatives}
Due to the presence of the ReLU activation in the auto-encoder (see Eq.\ (\ref{eq:autoencoder_definition})), the loss objective $\cL(\bw)$ in Eq.\ (\ref{eq:autoencoder_objective}) is not smooth everywhere since ReLU is not differentiable at $0$. The objective is in fact first-order differentiable everywhere due to the squaring, but not second-order differentiable at $\bw$ such that $\ip{\bw}{\ba_i} = 0$ for some $i \in [m]$. However, just as the ReLU function is one-sided differentiable everywhere, the loss objective has one-sided derivatives of all order everywhere. We will now introduce our notation for one-sided derivatives and related definitions.

Given a function $f : \R^n \to \R$ and a direction $\bv \in \R^n$, define the one-sided derivative of $f$ in the direction $v$ at a point $\bx \in \R^n$ as the following scalar quantity
\[
    D_{\bv}~f(\bx) := \lim\limits_{t \to 0^+} \frac{f(\bx + t\bv) - f(\bx)}{t}.
\]
We can define the second directional derivative analogously as
\[
D_{\bv}^2~ f(\bx) = \lim\limits_{t \to 0^+} \frac{D_{\bv}~f(\bx + t\bv) - D_{\bv}~f(\bx)}{t}.
\]
Note that if $f$ is first-order differentiable at $\bx$ then
\[
D_{\bv}~f(\bx) = \ip{\grad_{\bx}~f(\bx)}{\bv},
\]
and if $f$ is twice differentiable then 
\[
D_{\bv}^2~f(\bx) = \bv^\sT \grad^2_{\bx}~f(\bx) \bv.
\]
We will use these notions to define measures of sharpness which generalize the standard Hessian based measures.
\subsection{Generalized Sharpness}
In the literature, the two prevalent notions of sharpness of a function $\bx$ at a point $\bx$ are the maximum eigenvalue and the trace of the Hessian of $f$ at $\bx$. That is, if $f$ is twice-differentiable and $\bH(\bx) := \grad_{\bx}^2~f(\bx)$ has eigenvalues $\lambda_1 \geq \lambda_2 \geq \ldots \lambda_n$, then the two measures can be written as
\begin{enumerate}
    \item (Maximum Curvature) $\norm{\bH(\bx)}_2 = \lambda_1$,
    \item (Average Curvature) $\Tr(\bH(\bx)) = \sum_{i=1}^n \lambda_i$.
\end{enumerate}
The first quantity measures the curvature in the maximal direction. The second quantity can be seen to be a measure of average curvature over random directions since by a well-known identity
\[
    \Tr(\bH(\bx)) = \E_{\bv \sim \cN(0, \id)}~ \bv^\sT \bH(\bx) \bv.
\]

For both measures, large values indicate increased sharpness. We now introduce sharpness measures which generalize the previous ones, but are well-defined for functions with only one-sided derivatives
\begin{enumerate}
    \item (Maximum Curvature) $\norm{D^2 f(\bx)}_2 := \sup_{\norm{\bv} = 1} D^2_{\bv}~f(\bx)$,
    \item (Average Curvature) $\Tr(D^2 f(\bx)) := \E_{\bv \sim \cN(0, \id)}~ D_{\bv}^2~ f(\bx)$.
\end{enumerate}
From the previous section, we know that these measures are in fact generalizations, since if $f$ is differentiable then
\[
    \norm{D^2 f(\bx)}_2 = \norm{\bH(\bx)}_2,~~\Tr(D^2 f(\bx)) = \Tr(\bH(\bx)).
\]

\subsection{Sharpness at Global Minima}
Using the sharpness measures defined in the previous section, we now explicitly compute the sharpness of the loss objective $\cL(\bw)$ at the convergence points of GD and (C)SGD.

First note that we can rewrite the loss objective as
\begin{equation}\label{eq:objective_rewrite}
    \cL(\bw) = \frac{1}{m} \sum\limits_{i \in [m]} f_i(\bw)  + \text{const},~~f_i(\bw) = \phi^2(\ip{\bw}{\ba_i}) \cdot (\norm{\bw}^2/2 - 1).
\end{equation}
Indeed by expanding the square in Eq.\ (\ref{eq:autoencoder_objective})
\begin{align*}
    \cL(\bw) &= \frac{1}{2m} \sum\limits_{i \in [m]} \norm{\ba_i - \bw \phi(\sip{\bw}{\ba_i})}^2\\
    &= \frac{1}{m} \sum\limits_{i \in [m]} \frac{1}{2}\norm{\ba_i}^2 - \sip{\bw}{\ba_i} \phi(\sip{\bw}{\ba_i}) + \frac{1}{2} \norm{\bw}^2\phi^2(\ip{\bw}{\ba_i})\\
     &= \frac{1}{2} + \frac{1}{m} \sum\limits_{i \in [m]} - \phi(\sip{\bw}{\ba_i})^2 + \frac{1}{2} \norm{\bw}^2\phi^2(\ip{\bw}{\ba_i})\\
    &= \frac{1}{2} + \frac{1}{m} \sum\limits_{i \in [m]} \phi^2(\ip{\bw}{\ba_i}) \cdot (\norm{\bw}^2/2 - 1).
\end{align*}
Therefore we can write the second directional derivative of $\cL$ as
\begin{equation}\label{eq:sec_deriv_decomp}
    D^2_{\bv}~\cL(\bw) = \frac{1}{m} \sum_{i \in [m]} D^2_{\bv}~f_i(\bw).
\end{equation}
 Observe that if we define for $i \in [m]$,
\[
    g_i(\bw) = \ip{\bw}{\ba_i}^2\cdot (\norm{\bw}^2/2 - 1),
\]
then the second derivative of $f_i$ at $\bw$ is
\begin{equation}\label{eq:loss_obj_term_ddw}
    D^2_{\bv}~ f_i(\bw) = 
\begin{cases}
    D^2_{\bv}~g_i(\bw) & \text{if } \ip{\bw}{\ba_i} \geq 0 \text{ and } \ip{\bv}{\ba_i} \geq 0,\\
        0 & \text{otherwise}.
\end{cases}
\end{equation}
Since $g_i(\bw)$ is twice-differentiable we can compute the gradient and Hessian as
\begin{align}
    \grad_\bw~g_i(\bw) &= \ip{\bw}{\ba_i} (\norm{\bw}^2 - 2) \ba_i  + \ip{\bw}{\ba_i}^2 \bw,\nonumber\\
     \grad^2_\bw~g_i(\bw) &= (\norm{\bw}^2 - 2) \ba_i \ba_i^\sT + 2 \ip{\bw}{\ba_i}(\ba_i \bw^\sT + \bw \ba_i^\sT) + \ip{\bw}{\ba_i}^2 \cdot \id_n.\label{eq:gi_hessian}
\end{align}
For a set $\cS \subseteq [m]$, consider points 
\[
    \bw = \sum_{i \in \cS} c_i \ba_i~ \text{ such that } c_i > 0 \text{ for all } i \in \cS \text{ and } \sum_{i \in \cS} c_i^2 = 1.
\] 
We will be interested in such points because our convergence theorems show that GD converges to the point $\bw_{\GD}$ (see Eq.\ (\ref{eq:gd_critical_point})) for which $\cS = \cS^+$ and SGD converges to the point $\bw_{\SGD}$ (see Eq.\ (\ref{eq:sgd_critical_point})) for which $|\cS| = 1$.
\subsection{Proof of Theorem \ref{thm:sharpness}}
\paragraph{Computing Maximum Curvature} Note that if $\ip{\bw}{\ba_i} = 0$ and $\ip{\bv}{\ba_i} > 0$ then
\[
    \grad^2_\bw~g_i(\bw) = (\norm{\bw}^2 - 2) \ba_i \ba_i^\sT = -\ba_i \ba_i^\sT,
\]
which is negative semidefinite. Hence for $i \notin \cS$,  $D^2_{\bv} f_i(\bw) \geq D^2_{\bv} g_i(\bw)$. 
Therefore
if we are trying to maximize $D^2_{\bv} \cL(\bw)$ with respect to $\bv$, we can restrict our consideration to $\bv$ such that $\ip{\bv}{\ba_i} = 0$ if $i \notin \cS$. For such $\bv$, by Eq.\ (\ref{eq:sec_deriv_decomp}) and (\ref{eq:loss_obj_term_ddw})
\begin{align*}
    D^2_{\bv}~ \cL(\bw) &= \frac{1}{m} \sum\limits_{i \in \cS} D^2_{\bv}~ g_i(\bw)\\
    &= \frac{1}{m} \bv^\sT \qty(  \sum\limits_{i \in \cS} \grad_\bw^2~ g_i(\bw)) \bv\\
    &=  \frac{1}{m} \bv^\sT \qty( \sum\limits_{i \in \cS} 2 \ip{\bw}{\ba_i}(\ba_i \bw^\sT + \bw \ba_i^\sT))\bv + \frac{1}{m} \bv^\sT \qty( \sum\limits_{i \in \cS} -\ba_i\ba_i^\sT + \ip{\bw}{\ba_i}^2 \id_n)\bv \\
    &= \frac{1}{m} \bv^\sT \qty( \sum\limits_{i \in \cS} 2 \ip{\bw}{\ba_i}(\ba_i \bw^\sT + \bw \ba_i^\sT))\bv\\
    &=  \frac{4}{m} \sum\limits_{i \in \cS}  \ip{\bw}{\ba_i}\ip{\bv}{\ba_i}\ip{\bw}{\bv} \\
    &= \frac{4}{m} \ip{\bw}{\bv}^2.
\end{align*}
From this it is easy to see that $\bv = \bw$ is a maximizer and $\norm{D^2 \cL(\bw)}_2 = 4/m$.

\paragraph{Computing Average Curvature}

Now let us compute $\Tr(D^2 \cL(\bw))$.
Observe that from Eq.\ (\ref{eq:loss_obj_term_ddw})
\begin{align*}
    \Tr(D^2 f_i(\bw)) &= \Tr(D^2 g_i(\bw)) = \Tr(\grad^2_\bw~g_i(\bw)) && \text{if } \ip{\bw}{\ba_i} > 0\\
    \Tr(D^2 f_i(\bw)) &= \frac{1}{2}\Tr(D^2 g_i(\bw)) = \frac{1}{2}\Tr(\grad^2_\bw~g_i(\bw)) && \text{if } \ip{\bw}{\ba_i} = 0
\end{align*}
where the second line is due to the fact that $\Pr_{\bv \sim \cN(0, \id)}(\ip{\bv}{\ba_i} \leq 0) = 1/2$.
Note that 
\[
\Tr(\grad^2_\bw~g_i(\bw)) = (\norm{\bw}^2 - 2) + (n + 4)\ip{\bw}{\ba_i}^2.
\]
and that by Eq.\ (\ref{eq:sec_deriv_decomp}) and the linearity of $\Tr(D^2\; \bullet)$
\[
\Tr(D^2 \cL(\bw)) = \frac{1}{m} \sum\limits_{i \in [m]} \Tr(D^2 f_i(\bw)).
\]
 Therefore by Eq.\ (\ref{eq:loss_obj_term_ddw})
\begin{align*}
    \Tr(D^2 \cL(\bw)) &= \frac{1}{m} \qty[\sum\limits_{\ell \in \cS} \Tr(\grad^2_\bw~g_\ell(\bw)) + \sum\limits_{i \in [m] \setminus \cS} \Tr(\grad^2_\bw~g_i(\bw)) / 2]\\
    &= \frac{1}{m}\qty[\sum\limits_{\ell \in \cS^+} -1 + (n + 4)c_\ell^2 + \sum\limits_{i \in [m] \setminus \cS} (-1/2)]\\
    &= \frac{1}{m}[-|\cS| + (n+4) - (m - |\cS|) / 2] = \frac{2n + 8 - m - |\cS|}{2m}.
\end{align*}
Thus in particular we see that
\begin{align*}
    \Tr(D^2 \cL(\bw_{\GD})) &= \frac{2n + 8 - m - |\cS^+|}{2m},\\
    \Tr(D^2 \cL(\bw_{\SGD})) &= \frac{2n + 7 - m}{2m}.
\end{align*}
